\documentclass{article}

\usepackage{microtype}
\usepackage{graphicx}
\usepackage{subcaption}
\usepackage{amssymb}
\usepackage{booktabs} %

\usepackage{hyperref}

\usepackage[preprint]{icml2026}

\usepackage{amsmath,amsfonts,bm}

\newcommand{\procedure}{\ifmmode T \else $T$~\fi}
\newcommand{\target}{\ifmmode \mathcal{M} \else $\mathcal{M}$~\fi}
\newcommand{\explainer}{\ifmmode \mathcal{E} \else $\mathcal{E}$~\fi}

\def\eqref#1{equation~\ref{#1}}

\def\1{\bm{1}}

\DeclareMathAlphabet{\mathsfit}{\encodingdefault}{\sfdefault}{m}{sl}
\SetMathAlphabet{\mathsfit}{bold}{\encodingdefault}{\sfdefault}{bx}{n}

\DeclareMathOperator*{\argmax}{arg\,max}

\usepackage{hyperref}
\hypersetup{
    colorlinks=true,
    linkcolor=blue,
    citecolor=blue,
    urlcolor=blue,
}

\usepackage{url}
\usepackage{booktabs}
\usepackage{multirow}
\usepackage{xcolor}
\usepackage{tcolorbox}
\usepackage{textcomp}
\usepackage{dashrule}
\usepackage{bm}
\usepackage[T1]{fontenc}
\usepackage[utf8]{inputenc}  %
\usepackage{dsfont}
\usepackage{inconsolata}

\tcbuselibrary{listings,breakable}  %
\tcbset{listing engine=listings,colframe=black,colback=white,size=small}

\newcommand{\shadedsquare}[1]{%
  \begin{tcolorbox}[colframe=blue,colback=blue!10!white,boxrule=0.4pt,sharp corners,left=0.5em, right=0.5em, top=0.5em, bottom=0.5em,]
    #1
  \end{tcolorbox}%
}

\newenvironment{promptlist}{%
    \begin{list}{}{%
        \setlength{\leftmargin}{1em}%
        \setlength{\itemsep}{0.5em}%
        \setlength{\parsep}{0pt}%
    }%
}{%
    \end{list}%
}

\definecolor{placeholdercolor}{RGB}{150,0,130} %
\newcommand{\ph}[1]{\textcolor{placeholdercolor}{\textbf{\texttt{\detokenize{#1}}}}}
\definecolor{mathcolor}{RGB}{0,100,140}       %
\newcommand{\phm}[1]{{\mathcolor{placeholdercolor}{\bm{#1}}}}

\usepackage{xcolor}
\usepackage{listings}

\definecolor{promptframe}{gray}{0.40}
\definecolor{promptbg}{gray}{0.97} %

\lstdefinestyle{promptbox}{
  backgroundcolor=\color{promptbg},
  frame=single,
  rulecolor=\color{promptframe},
  framesep=8pt,
  framerule=0.4pt,
  basicstyle=\small\ttfamily,
  breaklines=true,
  breakatwhitespace=false,
  columns=fullflexible,
  keepspaces=true,
  showstringspaces=false,
  mathescape=true,
  moredelim=**[is][{\color{placeholdercolor}\ttfamily\bfseries}]{\{}{\}},
}

\usepackage[capitalize,noabbrev]{cleveref}

\usepackage[textsize=tiny]{todonotes}

\icmltitlerunning{Training Language Models to Explain Their Own Computations}

\begin{document}

\twocolumn[
  \icmltitle{Training Language Models to Explain Their Own Computations}

  \icmlsetsymbol{equal}{*}

  \begin{icmlauthorlist}
      \icmlauthor{Belinda Z. Li}{transluce,mit}
    \icmlauthor{Zifan Carl Guo}{mit}
    \icmlauthor{Vincent Huang}{transluce}
    \icmlauthor{Jacob Steinhardt}{transluce}
    \icmlauthor{Jacob Andreas}{transluce,mit}

  \end{icmlauthorlist}

\icmlaffiliation{transluce}{Transluce}
\icmlaffiliation{mit}{MIT CSAIL}
\icmlcorrespondingauthor{Belinda Z. Li}{bzl@mit.edu}

  \icmlkeywords{Machine Learning, ICML}

  \vskip 0.3in
]

\printAffiliationsAndNotice{}  %

\begin{abstract}
Can language models (LMs) learn to faithfully describe their internal computations? Are they better able to describe themselves than other models?
We study the extent to which LMs' privileged access to their own internals can be leveraged to produce new techniques for explaining their behavior.
Using existing interpretability techniques as a source of ground truth, we fine-tune LMs to generate natural language descriptions of
(1) the information encoded by LM features, (2) the causal structure of LMs' internal activations, and (3) the influence of specific input tokens on LM outputs.
When trained with only tens of thousands of 
example explanations,
explainer models exhibit non-trivial generalization to new queries.
This generalization appears partly attributable to explainer models' privileged access to their own internals:
fine-tuning a model to explain its \emph{own} computations generally works better than fine-tuning a different model (even if the explainer model is significantly more capable than the target).
Our results suggest not only that LMs can learn to reliably explain their internal computations, but that such explanations offer a scalable complement to existing interpretability methods.\footnote{Code available at \url{https://github.com/TransluceAI/introspective-interp}.}

\end{abstract}

\begin{figure*}[t]
    \centering
    \includegraphics[width=\linewidth, trim=10 650 10 20, clip]{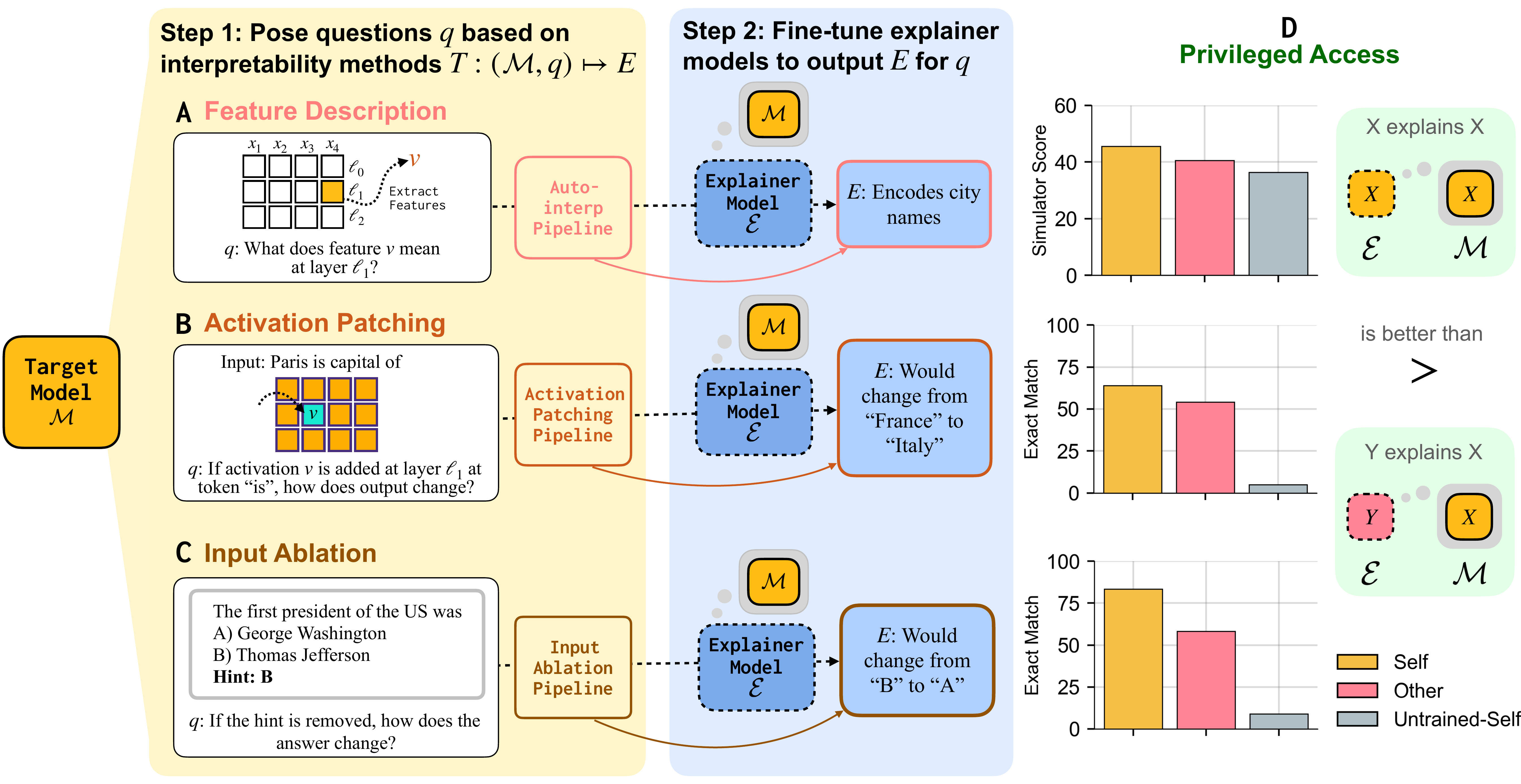}
    \vspace{-1.5em}
    \caption{Overview of our methods for training models to describe their own internal procedures. As a first step, we extract answers to three types of question about the target model using different interpretability methods: (A) descriptions of features of the target model's hidden representations via an auto-interp procedure, (B) explanations about what parts of the model affect the output via activation patching, (C) explanations about what parts of the input affect the output via input ablation. As a second step, we fine-tune three explainers to produce each of the three types of explanations. (D) After fine-tuning various explainer models on various target models, we find evidence of privileged access for all three explanation types, and use this to obtain data-efficient explainers.}
    \label{fig:teaser}
\end{figure*}

\section{Introduction}
\label{sec:intro}

When language models (LMs) are prompted to explain decisions, %
their outputs are often plausible~\citep{korbak2025chainthoughtmonitorabilitynew},
but may not bear any relationship to the %
computation that first gave rise to LMs' decisions.
Indeed, work using attribution, feature visualization, and mechanistic interpretability techniques \citep{Simonyan2013DeepIC,10.5555/3495724.3497168,templeton2024scaling} has shown that LMs' verbalized explanations sometimes systematically fail to describe the factors that determine their decisions \citep{turpin2023language,chen2025reasoningmodelsdontsay, barez2025cot}.

Can we fix this problem by training LMs so that their verbalized explanations faithfully describe their internal computations? Several recent papers \citep{binder2025looking,song2025privilegedselfaccessmattersintrospection,plunkett2025selfinterpretabilityllmscomplexinternal} have studied the related question of whether models can describe features of their own \emph{output distributions}, arguing that models might be able to leverage \textbf{privileged access} to their own internals to do so effectively.
In this paper, we are interested in an even deeper form of privileged access: whether models can learn to describe not only their outputs, but their \emph{internal} representations and mechanisms, and whether privileged self-access enables them to do so better than learned explanation-generating methods derived from other models. We state our main hypothesis as follows:

\shadedsquare{
\textbf{The Privileged Access Hypothesis} \\
Models trained to explain their own internal computations can do so more accurately than other models trained to explain them.
}

To explore this hypothesis, %
we fine-tune various \textit{explainer} models to predict three aspects of \textit{target} models' internal procedures:\footnote{This paper treats the \textit{target model} as fixed and only fine-tunes the \textit{explainer model}.
While this setup does not constitute \textit{self}-explanation in the strictest sense (since the explainer's output distribution changes), we believe it is more useful to explain pre-trained models than enforce this strict notion of self-consistency. We discuss the implications of this simplification in~\Cref{sec:discussion}. For the rest of this paper, we use the term ``self-explaining'' in this looser sense.} 
\begin{enumerate}
    \item Descriptions of internal features, specifically what inputs activate them (\textbf{feature descriptions}; \Cref{fig:teaser}A).%
    \item Outcomes of interventions to internal activations (\textbf{activation patching}; \Cref{fig:teaser}B).
    \item Descriptions of decision rules, specifically the important tokens for a decision (\textbf{input ablation}; \Cref{fig:teaser}C).
    This setting bears similarities to~\citet{binder2025looking} due to measuring self-access to the model's output distributions.
\end{enumerate}

Our experiments find evidence of privileged access---the target model is better explained by itself than by other explainer models, even when the other model is more capable (e.g.\ larger or instruction-tuned).
Diving deeper, we find:
\begin{enumerate}
    \item \textbf{LMs can be fine-tuned to self-explain}: As in past work \citep{li2025do}, we find that explanation capabilities are not always natively present in models, and must be acquired via fine-tuning. A small amount of fine-tuning enables LMs to faithfully describe their internal features, significantly outperforming zero-shot methods~\citep{kharlapenko2024self,chen2024selfie} and nearest-neighbor SAE 
    features~\citep{huben2024sparse}. %
    Furthermore, 
    a model’s ability to explain (its own or others’) features is correlated with the degree of similarity between the explainer and target models (\S\ref{sec:feature_explanations:results}).
    \item \textbf{Self-explaining is data-efficient}: 
    Self-explaining is approximately a hundred times more sample-efficient than a nearest neighbors baseline, achieving comparable results with only 0.8\% of the training data (\S\ref{sec:feature_explanations:scaling}).
    \item \textbf{Privileged access extends across tasks}: we 
    find patterns of privileged access in all three domains, including activation patching (\S\ref{sec:activation_patching}) and input ablation (\S\ref{sec:input_ablations}).
\end{enumerate}

These results suggest that %
even when off-the-shelf LMs cannot faithfully self-interpret, they can learn to do so through an objective that enforces consistency between their output and their internal procedures.
Our approach reframes interpretability as not only an external analysis problem, but as a capability that can be trained into LMs themeselves; by leveraging privileged access to internal computations, 
``introspective interpretability'' techniques offer an avenue towards scalable understanding of LM behavior.

\section{Methods}
\label{sec:method}
In this section we describe a general framework for training models to verbalize different aspects of their internal states, and then instantiate it with three specific explanation types.

\subsection{Definitions and Task}
\label{sec:background}

Abstractly, a neural \textbf{model} \target can be represented as a pipeline $x\to h\to y$ that maps \textbf{inputs} $x$ to \textbf{hidden representations} $h$ to \textbf{outputs} $y$. 
An \textbf{explanation} $E$ of the neural network %
may enable us to answer \textbf{questions} $q$ about what information in $x$ is represented in $h$ and how that information influences $y$.
For example, we may wish to answer a question $q=\textit{What inputs activate direction $v$ at layer $l$?}$ 
If $v$ activates on tokens \textit{Tokyo}, \textit{NYC}, and \textit{Paris}, then an appropriate explanation $E$ might be \textit{city names}. 
We are interested in whether LMs can learn to generate $E$ directly.

In this paper, we will train \textbf{explainer models} \explainer to produce explanations $E$ for various \textbf{target models} \target and types of questions $q$.
Abstractly, let $\procedure: (\target,q)\mapsto E$ be a (model-external) \textbf{explanation procedure}
that generates an explanation $E$ answering a question $q$ about model \target. 
We use \procedure as supervision to train an explainer model, by minimizing the cross-entropy loss on the explanation:
\begin{equation}
\label{eq:objective}
\mathcal{L}_\explainer
= 
- \mathbb{E}_{q} \left[
    \log p_\explainer(E = \procedure(\target, q) \mid q)
\right].
\end{equation}

Our experiments focus on explaining Transformer language models $\mathcal{M}$~\citep{grattafiori2024llama3herdmodels,yang2025qwen3technicalreport}, which consist of layers $\ell$ each containing attention and MLP blocks, and \textit{residual stream} representations $h_{\ell}$ in between each layer. We also use Transformer language models as explainers $\mathcal{E}$.
Throughout this paper, we use notation $x = (x_1,\cdots x_n)$ to refer to an input sequence with token positions $t\in [n]$. We use
$h_{(\ell,t)}(x)$
to denote the residual stream activation of \target at layer $\ell$ and token position $t$ for input $x$.

We will train explainer LMs to answer
three types of questions $q$, described below.

\subsection{Feature Descriptions: What does a model-internal feature represent?}
\label{sec:methods:feature_descriptions}
\begin{figure}
    \centering
    \includegraphics[width=\linewidth,trim=0 1100 250 0,clip]{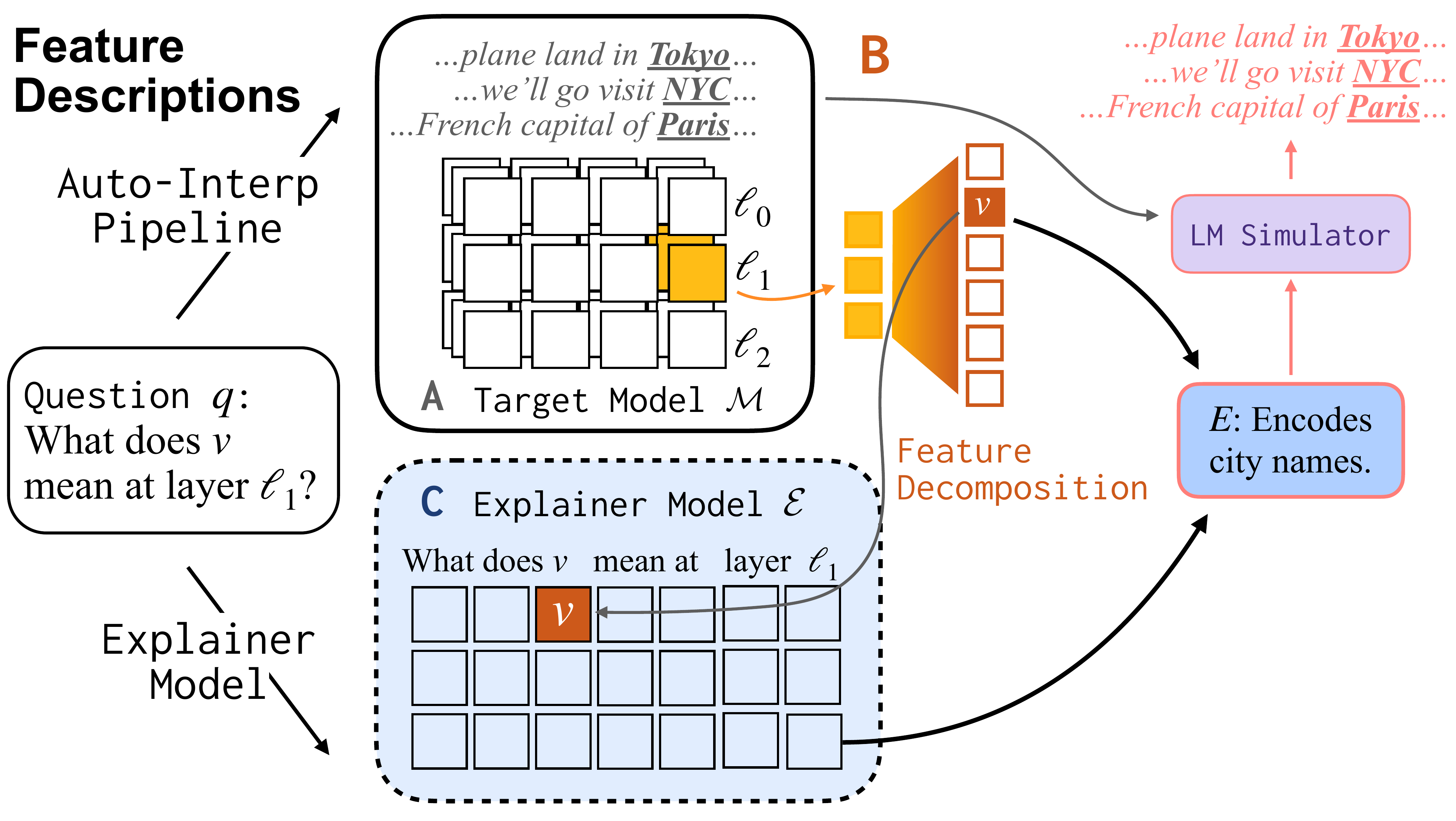}
    \vspace{-1em}
    \caption{Training an explainer to predict \textbf{feature descriptions}. (A) Run the target model $\mathcal{M}$ on many inputs. (B) For each vector $v$ at layer $\ell$, pick an explanation $E$ that best matches the context where $v$ is active, as judged by a LM simulator. (C) Train the explainer to output $E$ given the question $q$ about $v$ at layer $\ell$. The trained explainer generalizes well to both held-out and OOD $v$.}
    \label{fig:methods_features}
    \vspace{-1em}
\end{figure}

\paragraph{Question $q$.} For feature descriptions, our question $q$ takes form \textit{what types of inputs activate direction $v$ in the residual stream at layer $l$?} %

\paragraph{Interpretability procedure $T$.} We make use of automated interpretability methods~\citep{hernandez2022natural,bills2023language}, which generate natural-language descriptions $E$ of input tokens on which a given $v$ is highly active.

Formally, let
$a_v(x,\ell,t) = \langle h_{\ell,t}(x), v\rangle$
denote the ``activation'' of $v$ in the residual stream representation of token $x_t$ (i.e.\ the alignment between $h_{\ell,t}$ and $v$).
To obtain feature descriptions, we use %
a ``simulator'' language model to simulate how $v$ would activate on an input sequence $x$ if it had description $E$, outputting an ``expected'' activation pattern $\hat{a}(x, t, E)$.
See \Cref{fig:methods_features}B. We train the simulator following~\citet{choi2024automatic} using Neuronpedia SAE labels and FineWeb exemplars~\citep{penedo2024the}.

We then search for the explanation $E$ that maximizes correlation between simulated and true activations over inputs:
\begin{equation}
\label{eq:simulator_correlation}
\procedure_{\text{feat}}(\target,v,\ell)
= \operatorname*{arg\,max}_{E} 
\mathbb{E}_x\left[
\texttt{corr}_{t}\left(a_v(x,\ell,t),\, \hat{a}(x,t,E)\right) 
\right] 
\end{equation}
The prompts to encode $q$ and $E$ are given in~\Cref{app:prompts}.

\paragraph{Training the explainer.}
We take $v$ from the layer-$\ell$ residual stream of the target model
and pass it as a continuous token to the explainer by inserting $v$ at the embedding layer of the explainer model.%
\footnote{In early experiments, we also investigated explaining features $v$ from other components of the target model, but found that explainers struggled to reproduce these explanations. We hypothesize that we need to patch features into the same type of location (i.e. the embedding layer is part of the residual stream).}
See~\Cref{fig:methods_features}C.

For explainer models whose hidden dimension do not match the target model's, we introduce a \textbf{linear projection} $\Pi_\ell\in \mathbb{R}^{d_{\explainer}\times d_{\target}}$ from the target model hidden size $d_{\target}$ to the explainer model hidden size $ d_{\explainer}$, which is trained jointly with the rest of LM parameters. We learn a separate projection per layer $\ell$ of the target model. We train explainers and projections to optimize:
\begin{equation}
\label{eq:objective_features}
\hspace{-1em}
\boxed{
\mathcal{L}_\text{feat}
=
- \mathbb{E}_{v,\ell} \left[
    \log p_\explainer(E = \procedure_\text{feat}(\target, v,\ell) \mid \Pi_\ell\, v,\ell)
\right]
}
\end{equation}

\paragraph{Choice of directions $v$.} 
Our experiments train on Sparse Auto-Encoder (SAEs) features obtained from sparse dictionary learning approaches \citealp{huben2024sparse,bricken2023monosemanticity}),
but test generalization to full activations and activation differences. See details in~\Cref{sec:feature_explanations:evaluation:feature_types}.

\subsection{Activation Patching: What internal components affect the prediction?} 
\label{sec:methods:activation_patching}
\begin{figure}
    \centering
    \includegraphics[width=\linewidth,trim=0 950 30 20,clip]{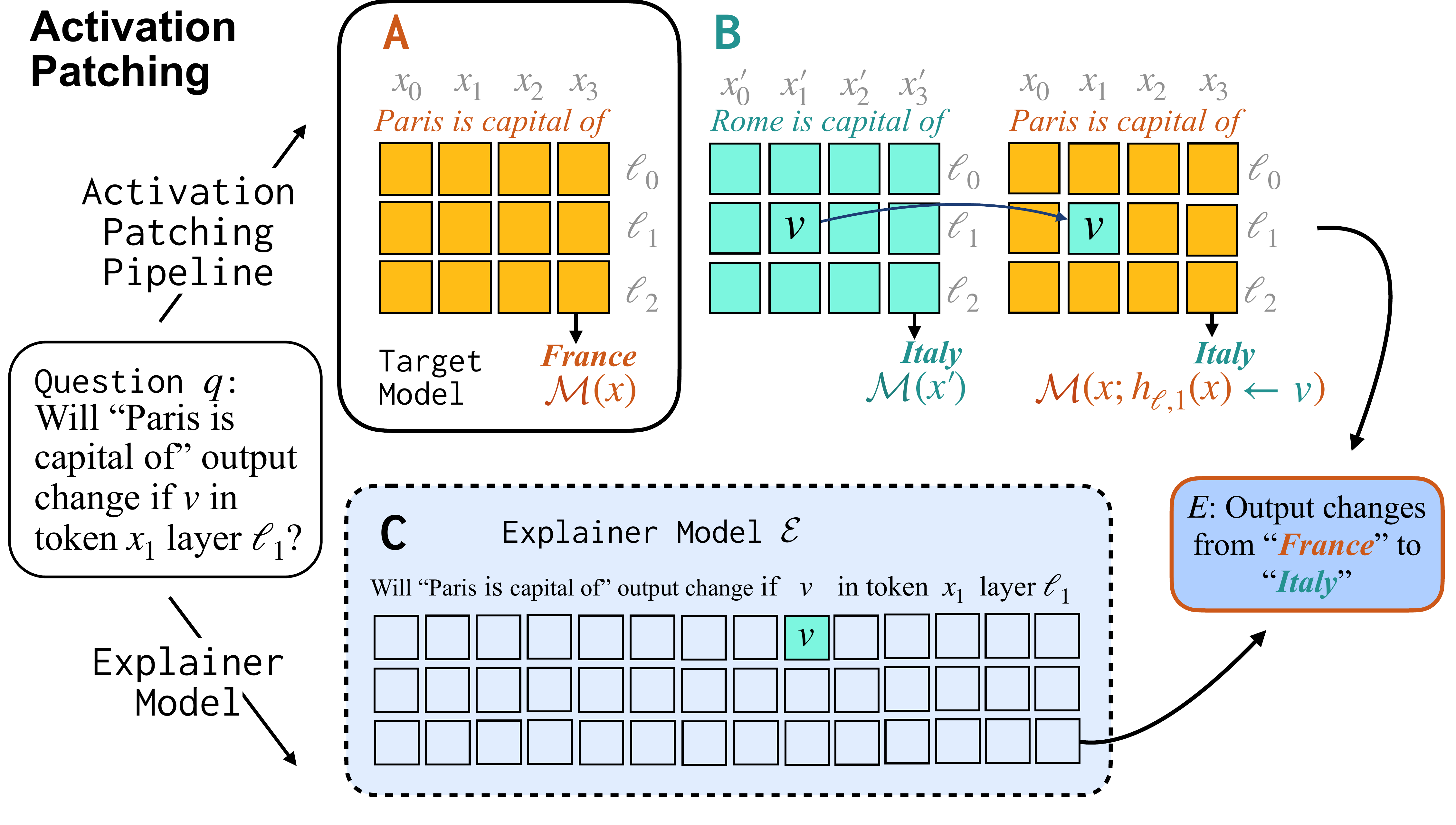}
    \vspace{-1em}
    \caption{Training an explainer to predict \textbf{activation patching outcomes}. (A) Run target model $\mathcal{M}$ on input $x$ to obtain $\mathcal{M}(x)$. (B) Perform activation patching by running $\mathcal{M}$ on a counterfactual input $x'$---in this example, patching in vector $v$ from token $x_1$ and layer $\ell_1$ of the counterfactual run into the original run---and then construct an explanation $E$ about how the resulting prediction changes. (C) Train an explainer model to answer $E$ when given a question $q$ about the patching procedures.}
    \label{fig:methods_actpatch}
    \vspace{-0.5em}
\end{figure}
\paragraph{Question $q$.} We ask which hidden components are \textit{causally important} for a model's output~\citep{meng2022locating,zhang2024towards}.
Specifically, the question $q$ is \emph{on input $x$, (how) does changing the activation at layer $\ell$ and token $x_t$ affect the output?} An example is shown in~\Cref{fig:methods_actpatch}. 

\paragraph{Interpretability Procedure $T$.}
We can answer questions of this form through activation patching and circuit tracing methods.
Given an input $x$ (\emph{Paris is the capital of} in \Cref{fig:methods_actpatch}), an activation patching experiment typically constructs an alternative input $x'$ (\emph{Rome is the capital of}). %
We then run a forward pass of $\mathcal{M}$ on $x$, but replace the activation $h_{\ell,t}(x)$ with the corresponding activation $h_{\ell,t}(x')$:
\begin{equation}\target\left(x;\, h_{\ell,t}(x) \leftarrow h_{\ell,t}(x')\right) ~ ,\end{equation}
and we measure the change in the model's output:
\begin{equation}
\label{eq:act_patch}
\begin{aligned}
T_{\text{patch}}(\target,q)
&= \procedure_{\text{patch}}(\target,x,\ell,t) \\
&= d\Big(
    \target(x),\;
    \target\big(
        x;\,
        h_{\ell,t}(x) \leftarrow h_{\ell,t}(x')
    \big)
\Big).
\end{aligned}
\end{equation}

where $d(\cdot,\cdot)$ is some measure of the difference between the two predictions.
This process is illustrated in~\Cref{fig:methods_actpatch}B.\footnote{
To ensure sufficient samples where the model's output changes
We divide the target model's layers into four blocks, and 
perform activation patching over
one block of layers at a time
We %
aggregate the representations over the block
$v=\texttt{avg}_{\ell_k}\left(h_{\ell_k,t}(x')\right)$
before inserting
into all corresponding layers of the original input, allowing us to specify only a single continuous token to the explainer.
}

\paragraph{Training the explainer.}
Rather than directly predicting the numerical score in \Cref{eq:act_patch}, we train explainers $\explainer$ to predict
(1) \textit{whether} the model output changes under patching,
and (2) the \emph{content} of the model output after patching. %
Similarly to~\Cref{sec:methods:feature_descriptions}, we insert $v$ as a continuous token in the embedding layer of the explainer model. See~\Cref{fig:methods_actpatch}C.

Finally, we optimize:
\begin{equation}
\label{eq:objective_patch}
q \triangleq (x, t, \ell_{1\cdots i}, v).
\end{equation}
\begin{equation}
\boxed{
\mathcal{L}_{\text{patch}}
=
- \mathbb{E}_{x,x',q}
\Big[
\log p_{\explainer}
\big(
E = \procedure_{\text{patch}}(\target,q)
\mid q
\big)
\Big]
}
\end{equation}

The exact prompt and output templates are in~\Cref{app:prompts}.

\subsection{Input Ablation: What parts of the input matter?}
\label{sec:methods:input_ablations}
\begin{figure}[t!]
    \centering
    \includegraphics[width=\linewidth,trim=10 1300 70 10,clip]{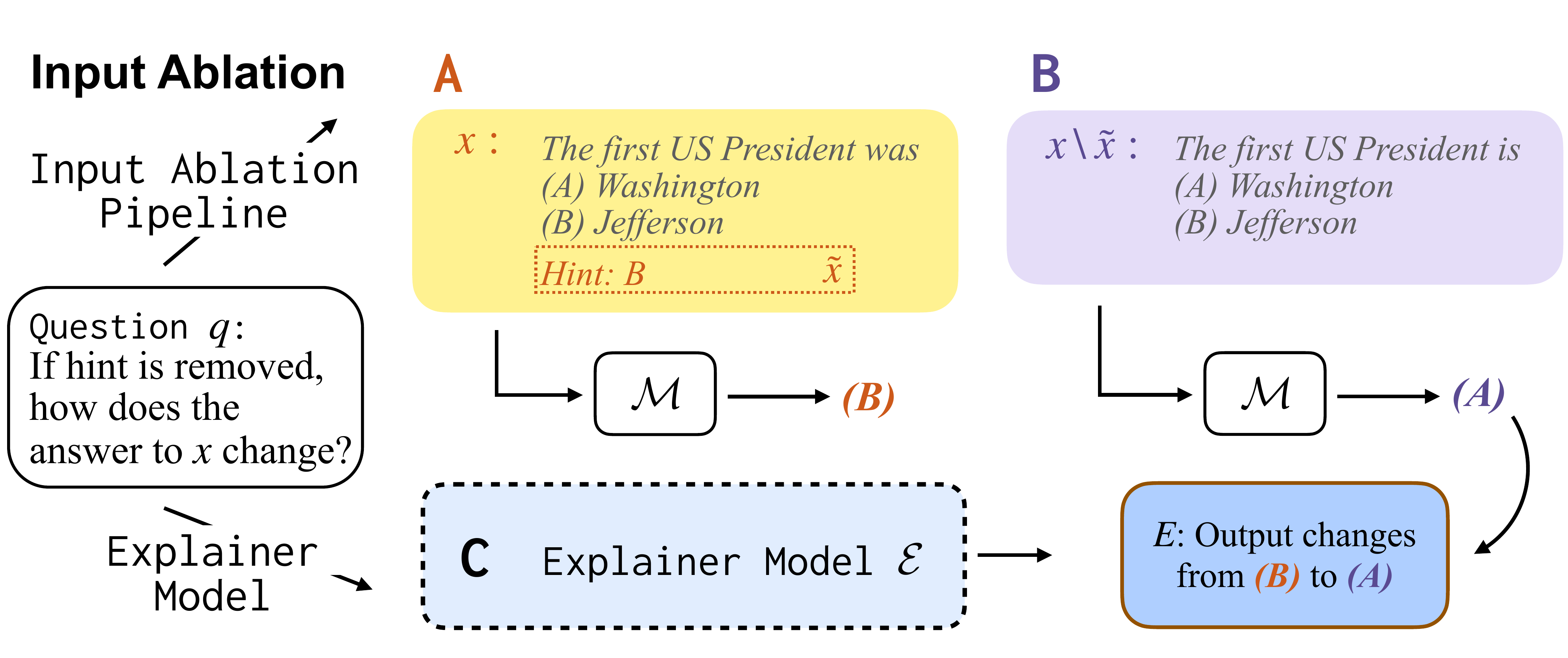}
    \vspace{-1em}
    \caption{Training an explainer to predict \textbf{input ablation outcomes}. (A) Run a target model $\mathcal{M}$ on an input $x$. (B) Identify whether input component $\tilde{x}\subset x$ affected its output by rerunning $\mathcal{M}$ on $x\backslash\tilde{x}$. (C) Train an explainer model to answer question $q$ about how the output changes when $\tilde{x}$ is removed from $x$.}
    \label{fig:methods_input}
    \vspace{-0.5em}
\end{figure}
\paragraph{Question $q$.}
For input ablation, we ask which input tokens most influence the model's prediction~\citep{chang-etal-2025-jopa,vafa-etal-2021-rationales}, which we can identify by ablating subsets $\tilde{x}$ of the input $x$. Specifically, $q$ takes form
\textit{if $\tilde{x}$ is removed, (how) does the answer to $x$ change?} %
See~\Cref{fig:methods_input}.

\paragraph{Interpretability Procedure $T$.}
Input ablation measures how much the target model $\mathcal{M}$'s output changes when 
we take an input $x$ (e.g.\ \emph{The first US president...} in \Cref{fig:methods_input}) and remove a subset of tokens $\tilde{x}\subseteq x$ (e.g.\ \emph{Hint: B}): 
\begin{equation}
\label{eq:input_ablate}
\procedure_{\text{input}}(\target,x,\tilde{x})
= d\left(\target(x),\, \target(x\backslash\tilde{x})\right),
\end{equation}
where $d$ measures difference in the model's outputs.
This process is illustrated in~\Cref{fig:methods_input}B.

\paragraph{Training the explainer.}
\explainer %
is asked to predict (1) \textit{whether} removing $\tilde{x}$ would change the $\mathcal{M}$'s answer,
and (2) the \textit{content} of $\mathcal{M}$'s new output without $\tilde{x}$.
See~\Cref{fig:methods_input}C. Formally, the objective we optimize is:
\begin{equation}
\label{eq:objective_input}
\hspace{-1em}
\boxed{
\mathcal{L}_\text{input}
=
- \mathbb{E}_{x,\tilde{x}} \left[
    \log p_\explainer(E = \procedure_\text{input}(\target, x,\tilde{x}) \mid x,\tilde{x})
\right]
}
\end{equation}
Full prompt and output templates are in~\Cref{app:prompts}.

\section{Privileged Access Improves Explanations}
\label{sec:feature_explanations}
In this section, we show that (1) models can be fine-tuned to generate feature descriptions and generalize to new features and feature bases, and (2) this capability relies on privileged access.
We focus on feature description (\S\ref{sec:methods:feature_descriptions}), 
and study additional tasks (patching, input ablation) in~\S\ref{sec:generalization}.

\label{sec:feature_explanations:experiments}
\subsection{Models and Training Data} 
\paragraph{Models.}
We use Llama-3.1-8B~\citep{grattafiori2024llama3herdmodels} as the target model,\footnote{We run additional experiments with Gemma 2~\citep{gemmateam2024gemma2improvingopen} as target model, see~\Cref{app:feature_explanations:gemma}.} and train five different explainer models (Llama-3.1-8B, Llama-3-8B, Llama-3.1-8B-Instruct, Llama-3.1-70B, Qwen3-8B;~\citealp{yang2025qwen3technicalreport}).\footnote{Details can be found in~\Cref{app:feature_explanations:models}.} 
For Llama-3.1-8B, we report results with both LoRA~\citep{hu2022lora} and full-parameter fine-tuning. For Llama-3.1-70B we only report LoRA; for other models we only report full  fine-tuning.

\paragraph{Initializing $\Pi_\ell$.} For models that require projection layers $\Pi_{\ell}$ due to hidden-size mismatch (Qwen3-8B, Llama-3.1-70B), we \textit{randomly initialize} $\Pi_\ell$ and train them with the rest of the parameters.\footnote{For LoRA training, these layers are included in the set of parameters trained via low-rank updates.} 
For Llama-3.1-70B, we also include results where $\Pi_\ell$ was \textit{pre-trained} on $\min_{\Pi_\ell}\|h^\explainer_{\ell, t} - \Pi_{\ell} \cdot h^\target_{\ell, t} \|_2$ (distance between explainer and projected target activations) on FineWeb samples~\citep{penedo2024the}, 

\paragraph{Data.} We train on residual stream SAE features of Llama-3.1-8B target models from Llama-Scope~\citep{he2024llamascope}.
We use the Llama-3.1-8B-LXR-32x features, which maps the residual stream after each layer to 131K features (32 times larger than the residual hidden dimension). Ground-truth explanations $E$ for each feature were obtained from Neuronpedia~\citep{neuronpedia}.

\label{sec:feature_explanations:evaluation:targets_explainers}

\subsection{Evaluation Set and Metrics}
\label{sec:feature_explanations:evaluation:feature_types}
We train on a subset of SAE features, then evaluate generalization in- and out-of-distribution on three types of features:

\textbf{Held-out SAE features (SAE):} We hold out a subset of 1550 SAE features (50 per layer) for testing.

\textbf{Full Activations (ACT):} We test out-of-distribution generalization to full residual stream activations $h_{(\ell,t)}(x)$ on inputs $x$ sampled from FineWeb~\citep{penedo2024the}.

\textbf{Activation Differences ($\Delta$ACT):} Differences between counterfactual pairs of inputs allow us to isolate the effect of a single pointwise change. We create counterfactual pairs $(x,x')$ from the Counterfact dataset~\citep{meng2022locating} using the procedure described in~\Cref{sec:methods:activation_patching} and extract activation differences $v = h_{\ell,t}(x) - h_{\ell,t}(x')$. \\

\label{sec:feature_explanations:evaluation:metrics}
To measure explainer performance on each type of feature, we use two metrics. \textbf{(1)} For SAE features, gold descriptions exist, so an LM judge can assess the
similarity of predictions to the gold description, on a scale of 0 to 1 in increments of 0.25. The LM judge has 81.25\% human agreement pe~\Cref{app:feature_explanations:llm_judge}.
\textbf{(2)} We also use \textit{simulator score}~(\Cref{eq:simulator_correlation}), measuring the Pearson correlation between true and predicted activations based on $\explainer(v)$ for each sample $x$.

\subsection{Baselines}
\label{sec:feature_explanations:evaluation:baselines}

\paragraph{Nearest Neighbors.} %
To evaluate whether explainer models exhibit nontrivial generalization to unseen features, we compare to a nearest-neighbor baseline that, given a new feature or representation, finds the single most similar feature from the training set, and outputs its associated description:
    $$\explainer_{\text{NN}}(v) %
    = D\!\left(
        \argmax_{v_i \in S_{\text{train}}} 
        \langle v_i, v \rangle
      \right).$$
where $S_{\text{train}}$ denotes the set of features used for training.
This gives us an estimate of roughly how well the explainer can do if it simply learned to memorize and interpolate samples from the training data.\footnote{We also perform \textit{layer-wise} nearest neighbors in~\Cref{app:feature_explanations:nearest_neighbors_layerwise}.}

\begin{table}[t!]
\centering
\footnotesize
\setlength{\tabcolsep}{2pt}
\begin{tabular}{@{}lcccc@{}}
\toprule
& \multicolumn{2}{c}{SAE} & & \\
\cmidrule(lr){2-3}
Explainer & LM Judge & Sim. & {ACT} & {$\Delta$ACT} \\
\midrule
Llama-3.1-8B
& \textbf{76.2$_{\pm 0.9}$} & \textbf{45.1$_{\pm 0.7}$} & \textbf{49.7$_{\pm 0.6}$} & {32.0$_{\pm 0.8}$} \\
Llama-3.1-8B (LoRA)
& \textbf{76.0$_{\pm 0.9}$} & \textbf{45.6$_{\pm 0.8}$} & \textbf{50.6$_{\pm 0.6}$} & \textbf{34.0$_{\pm 0.8}$} \\
Llama-3-8B
& \textbf{77.0$_{\pm 0.9}$} & \textbf{44.6$_{\pm 0.7}$} & {49.3$_{\pm 0.6}$} & {32.4$_{\pm 0.7}$} \\
Llama-3.1-8B-Inst. & \textbf{77.1$_{\pm 0.8}$} & 42.7$_{\pm 0.7}$ & 46.9$_{\pm 0.7}$ & 29.9$_{\pm 0.7}$ \\
Qwen3-8B & 70.3$_{\pm 0.9}$ & 40.6$_{\pm 0.8}$ & 21.1$_{\pm 0.7}$ & 12.3$_{0.4}$ \\
Llama-3.1-70B (LoRA)\\ 
~~random proj.
& 63.9$_{\pm 1.0}$ & 39.5$_{\pm 0.8}$ & 12.6$_{\pm 0.4}$ & 12.2$_{\pm 0.4}$ \\
~~pre-trained proj. 
& 74.1$_{\pm 0.9}$ & \textbf{45.2$_{\pm 0.7}$} & 33.8$_{\pm 0.7}$ & 20.6$_{\pm 0.6}$ \\
\midrule
Nearest Neighbors & 58.5$_{\pm 1.0}$ & 33.7$_{\pm 0.8}$ & 38.9$_{\pm 0.7}$ & 18.5$_{\pm 0.6}$ \\
SelfIE Best of 5 & 40.1$_{\pm 1.0}$ & 36.4$_{\pm 0.8}$ & 43.3$_{\pm 0.6}$ & 21.0$_{\pm 0.6}$ \\
\midrule
Gold SAE Labels & 100$_{\pm 0.0}$ & 43.3$_{\pm 0.8}$ & - & - \\
\bottomrule
\end{tabular}

\caption{\textbf{A target model's features are best explained by itself and its variants.}
Comparison of explainer methods on Llama-3.1-8B residual SAE features, including trained models, nearest neighbor, SelFIE, and gold SAE label baselines.
We report LM judge and simulator scores (mean $\pm$ SE) for held-out SAE features, and simulator scores for out-of-domain activations (ACT) and differences ($\Delta$ACT), all scaled to 100.
\textbf{Bold} means not significantly different from best (paired $t$-test, $p\ge 0.05$). Our methods can even beat gold SAE labels as scored by a simulator.
\textbf{Activation alignment improves explainer performance}: pre-training a projection to align features improves Llama-3.1-70B's performance.
}
\label{tab:feature_description_results}
\vspace{-2em}
\end{table}

\paragraph{SelfIE.} Following SelfIE~\citep{chen2024selfie}, we directly patch $v$ into the embedding layer of a non-finetuned Llama-3.1-8B model,\footnote{The original SelfIE paper patches representations into the third layer, but prior work has found that patching into the first few layers work about equivalently well (which we also find). Thus, we patch every activation into the embedding layer.} and prompt it to define the feature. We use standard prompts found to work well for self-explanations~\citep{kharlapenko2024self}.\footnote{SelfIE prompts can be found in~\Cref{app:untrained_prompts:selfie}.} Because SelfIE is sensitive to the scale of $v$ when it is inserted~\citep{kharlapenko2024self}, we report results on the \textit{best} explanation across 5 possible scales ($v$, $5v$, $10v$, $25v$, and $50v$), %
representing an upper bound on SelfIE performance.

\paragraph{Gold SAE labels.} For held-out SAE features, we evaluate their \textbf{gold Neuronpedia labels} with the simulator.

\subsection{Results}
\label{sec:feature_explanations:results}
Results can be found in~\Cref{tab:feature_description_results}. 
We observe:

\textbf{Trained explainers are effective.} For every feature type and metric, training Llama-3.1-8B to explain itself outperforms all baselines by a significant margin. 
This even includes gold SAE labels
due to noise in the ground-truth feature labels themselves (see error analysis in~\Cref{app:error_analysis}).

\textbf{Alignment between activations improves verbalization capability.} Among all trained models, the Llama-3.1-8B and Llama-3-8B models consistently outperform Llama-3.1-8B-Instruct by a few points and significantly outperform Qwen3-8B and Llama-3.1-70B, despite Llama-3.1-70B's larger size. We posit that activation similarity between the explainer and target model predicts explainer performance. 

We confirm this correlation visually by
plotting explainer-target activation similarity vs. explainer performance %
in ~\Cref{app:feature_explanations:alignment}. 
the explanation capabilities of Llama-3.1-70b with the \textit{pre-trained} projection and \textit{randomly initialized} projection serve as a proxy for maximally and minimally aligned activation, respectively, given the model. 
As reported in~\Cref{tab:feature_description_results}, we can recover a significant fraction of performance when we start training from the pre-trained projection, performing 14\% better on SAE explanations, 2.7 times better on real activations, and 1.7 times better on activation differences. 
We conclude that:
\shadedsquare{%
For feature descriptions, privileged access improves model explanations: models are better explained by fine-tuned versions of themselves than by other models.
}

\begin{figure*}
    \centering
    \includegraphics[trim={0 5 0 25},clip,width=\linewidth]{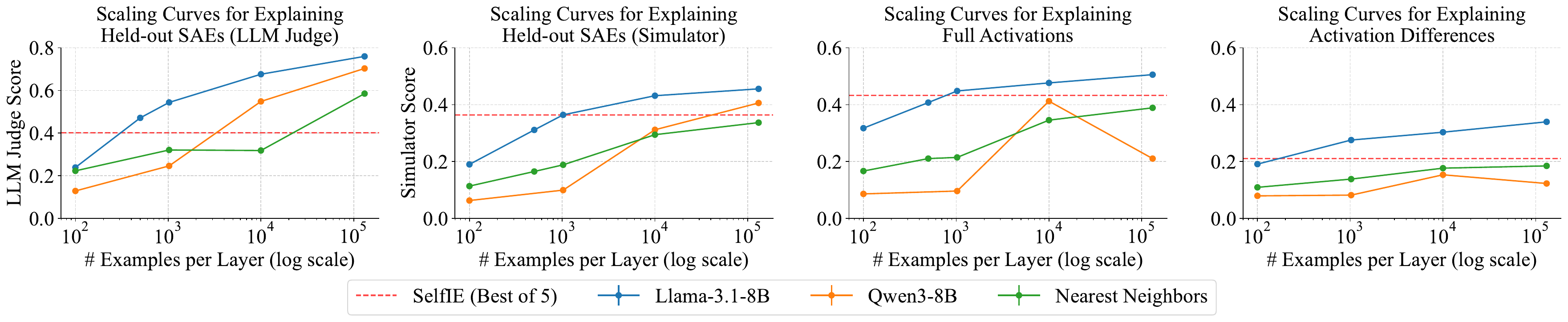}
    \caption{\textbf{Training self-as-explainer is more data-efficient than alternatives.} Scaling curves of training samples per layer vs. explanation quality (as measured either by LM judge or simulator) on three types of features (SAE, full activations, activation differences) show that matching explainer to target (Llama-3.1-8B) is more data-efficient than using a different model (Qwen) or nearest neighbors, especially in the low-data regime. The latter requires more samples per layer to outperform an untrained SelfIE baseline. Note that we aggregate 5 attempts of SelfIE with different strength, so the graph shows an upper bound of SelfIE performance. }
    \label{fig:scaling_simulator}
\end{figure*}

\section{Privileged Access Confers Data-Efficiency}
\label{sec:feature_explanations:scaling}

To answer where and why privileged access may be useful, we investigate whether it confers data-efficiency advantages over alternatives.
We train Llama-3.1-8B, Qwen3-8B, and the nearest neighbor baseline on subsets of the training data of various sizes and plot scaling curves against SelfIE baseline in~\Cref{fig:scaling_simulator}.

In general, self-explanation 
is more data-efficient than training another explainer model or using nearest neighbors.
At just 0.8\% of all training features (1024 samples per layer), 
Llama reaches 71\%, 80\%, 89\%, and 81\% of its end-performance on two metrics for held-out SAEs and simulator score for ACT and $\Delta$ACT respectively; Qwen reaches 35\%, 24\%, 46\%, and 66\%, and nearest neighbors reaches 55\%, 56\%, 55\%, and 75\%.
As Llama’s end-performance is already higher than Qwen and nearest neighbors, 
this gap only widens in low-data regimes. 
Thus,
\shadedsquare{%
Training a model to generate descriptions of its own features is data-efficient, particularly compared to training other models to describe the target model's features.
}

This data-efficiency underscores the practical utility of self-explanation: annotating feature descriptions is expensive, and self-explanation training allows us to recover most of the performance of other explanation techniques with orders of magnitude less data.

\section{Generalization Across Tasks}
\label{sec:generalization}
We check if self-explanation and privileged access generalize to other domains by training models to explain outcomes of activation patching (\S\ref{sec:methods:activation_patching}) and input ablation (\S\ref{sec:methods:input_ablations}).

\paragraph{Models.} Across both tasks, we train two explainer models, Qwen3-8B and Llama-3.1-8B, to explain Qwen3-8B. We also report results with Llama-3.1-8B as the target model for activation patching and input ablations in~\Cref{app:activation_patching:llama_target,app:input_ablation:full_results} respectively.

\paragraph{Metrics.}
\label{sec:act_patch:metrics}
Recall from~\Cref{sec:methods:activation_patching,sec:methods:input_ablations} that the explainer output for both activation patching and input ablations typically includes two parts: whether the target output would changed under intervention (\textit{changes} / \textit{remains unchanged}), and the content of the change (\textit{to France}). Based on this, we report three different types of metrics:
\begin{enumerate}
    \item Has-Changed F1: Correctness when predicting whether 
    the target's output would change under intervention. We report the Macro F1 scores over the ``changed'' and ``unchanged'' classes.
    \item Content Match: Correctness of predictions about the \textit{content} of target model output under intervention.
    \item Exact Match: Exact match accuracy between generated explanation and ground-truth explanation. Requires the explainer to correctly predict both parts.
\end{enumerate}

\paragraph{Baselines.}
To measure the effect of fine-tuning, we prompt Qwen3-8B (without any explanation training) to generate explanations of its own activation patching and input ablation procedures.
The full prompts 
for untrained activation patching are in~\Cref{app:untrained_prompts:activation_patching}. and for untrained input ablations are in~\Cref{app:untrained_prompts:input_ablations}. 
These baselines are analogous to the SelfIE baseline in~\Cref{sec:feature_explanations:evaluation:baselines}.

\begin{table*}[t]
\centering
\small
\resizebox{\textwidth}{!}{%
\begin{tabular}{llcccccc}
\toprule
& & \multicolumn{3}{c}{Activation Patching} & \multicolumn{3}{c}{Input Ablation} \\
\cmidrule(lr){3-5} \cmidrule(lr){6-8}
Target & Explainer & Exact Match & Has-Changed F1 & Content Match & Exact Match & Has-Changed F1 & Content Match \\
\midrule
\multirow{3}{*}{Qwen3-8B} & Qwen3-8B & \textbf{64.0$_{\pm 0.4}$} & \textbf{80.2$_{\pm 0.3}$} & \textbf{71.0$_{\pm 0.4}$} & \textbf{83.4$_{\pm 1.0}$} & \textbf{87.0$_{\pm 1.1}$} & \textbf{90.6$_{\pm 0.8}$} \\
& Llama-3.1-8B & 54.1$_{\pm 0.4}$ & 73.9$_{\pm 0.4}$ & 65.4$_{\pm 0.4}$ & 58.1$_{\pm 1.3}$ & 70.5$_{\pm 0.0}$ & 71.6$_{\pm 1.2}$ \\
& Qwen3-8B (Untrained) & 5.02$_{\pm 0.2}$ & 24.4$_{\pm 0.8}$ & 18.8$_{\pm 0.3}$ & 8.9$_{\pm 1.5}$ & 44.4$_{\pm 1.6}$ & 35.3$_{\pm 2.5}$ \\
\bottomrule
\end{tabular}%
}
\caption{\textbf{A target model's intervention outcomes are best predicted by itself, shown through activation patching and input ablation.}
We train Qwen3-8B to verbalize its own intervention outcomes, and compare against training a different model (Llama-3.1-8B), and also against an untrained version of Qwen3-8B.
Scores (mean $\pm$ standard error) are reported for three metrics---exact match, has-changed F1, and content prediction.
For each task and metric, \textbf{bold} indicates no significant difference from the best entry (paired $t$-test, $p\geq0.05$).
Results with Llama-3.1-8B as the target model can be found in~\Cref{tab:act_patch_results_full,tab:llama_target_input_ablation}.
}
\label{tab:act_patch_results}
\end{table*}

\subsection{Activation Patching: Training LMs to Predict the Outcomes of Activation Interventions}
\label{sec:activation_patching}

We train models to explain outcomes of
activation patching, which perturbs a model component and examines whether it changes the model output, described in detail in~\Cref{sec:methods:activation_patching}.

\paragraph{Training Data.} 
We use the CounterFact dataset~\citep{meng2022locating} as a source of counterfactual input pairs $x,x'$, which consists of a corpus of factual subject--relation--object sentences like \textit{Paris (subject) is the capital of (relation) France (object)}. We prompt the model with the subject and relation %
and ask the model to predict the object
. We then identify a counterfactual sentence $x'$ with the same relation but differing subject and object (\textit{Rome is the capital of}) and measure whether the most likely LM prediction changes (e.g.~to \textit{Italy}).
We perform sampling to ensure an even split between changed and unchanged predictions, at each token position and layer chunk. See~\Cref{app:training_data:activation_patching} for details.

\paragraph{Results} for activation patching can be found in~\Cref{tab:act_patch_results}. 
The privileged access hypothesis is supported: Qwen3-8B is best at explaining Qwen3-8B. All trained explainers perform better than their untrained versions, which generally guess the has-changed value randomly.
We also include a set of ablation experiments in~\Cref{app:activation_patching:ablation_analysis} to isolate which input component(s) the explainer learned to condition on.

\subsection{Input Ablations: Training LMs to Describe their Decision Rules}
\label{sec:input_ablations}
Using the methods described in~\Cref{sec:methods:input_ablations}, we train models to explain how withholding or including parts of target models' \textit{input} would affect the output prediction. 

\paragraph{Training Data.} 
Following recent work used to study faithful chain-of-thought~\citep{chen2025reasoningmodelsdontsay}, we give \target inputs in the form of a multiple-choice question $c$ with an injected hint $\tilde{x}$, i.e. $x=[c,\tilde{x}]$, and train the explainer to predict whether \target's most likely next token will change to $\tilde{x}$.
Previously,~\citet{chen2025reasoningmodelsdontsay} found that models fail to verbalize $\tilde{x}$ in their chain-of-thought, despite $\tilde{x}$ being critical to its prediction ($\target([c,\tilde{x}])\neq\target(c)$). Thus, we investigate whether models can be trained to detect \textit{when and how} their prediction will change depending on $\tilde{x}$'s presence.

We use MMLU~\citep{hendryckstest2021} as the source of multiple-choice reasoning question, and inject hints of the form $\tilde{x}=$ ``\textit{Hint: A}'' to the end of the question. We randomly select $\tilde{x}$ as one of the four multiple-choice options. To make the has-changed prediction nontrivial, we construct the hint prompt such that we have a roughly equal number of samples that change and do not change according to the hint. See~\Cref{app:training_data:input_ablation} for details.

\paragraph{Results} for input ablation are also shown in~\Cref{tab:act_patch_results}. 
We find similar evidence for \textit{privileged access}, where Qwen3-8B is best explained by itself than other models. 
Finally, we examine untrained Qwen's predictions and find that it a has-changed value of \verb|True| only 8.6\% of the time, and most of its content match comes from correctly predicting the unchanged answers (96.4\% content match) compared to its changed answers (10.3\% content match).
This is consistent with findings on reasoning models in prior work:~\citet{chen2025reasoningmodelsdontsay} finds that no pre-trained LM would ever report using a hint.
This indicates that explicit fine-tuning is essential to elicit faithful explanations of decision rules. 
Taken together, results on these two tasks show that:
\shadedsquare{%
Privileged access holds across tasks.
}

\section{Related Work}
\label{sec:related_work}
\subsection{Chain-of-Thought Faithfulness}
Language models can be asked to verbalize their thought processes through chain-of-thought, which offers a way for external monitoring~\citep{korbak2025chainthoughtmonitorabilitynew,baker2025monitoringreasoningmodelsmisbehavior}, but
prior work has found that these verbalizations can be unfaithful to their true decision-making processes~\citep{turpin2023language,lanham2023measuringfaithfulnesschainofthoughtreasoning,barez2025cot,chen2025reasoningmodelsdontsay}. This directly inspired our hint ablation task setup, where our work attempted to remedy unfaithfulness through training directly on the models' own decision rules.

\subsection{Mechanistic Interpretability Methods}
Mechanistic interpretability sets out to (1) describe individual features and (2) connect them to construct a causal circuit that performs a certain task. While many researchers have found success doing so manually~\citep{gurnee2024universal,wang2023interpretability,nanda2023progressmeasuresgrokkingmechanistic}, %
these methods have been challenging to scale and generalize to new models and tasks~\citep{sharkey2025open}.
To address this, automated feature description pipelines~\citep{hernandez2022natural,bills2023language,choi2024automatic,paulo2025automatically} and circuit discovery techniques~\citep{conmy2023automated,syed-etal-2024-attribution,hanna2024have,hsu2025efficient} have been introduced, though these often require substantial compute. Researchers are continuing to actively develop scalable interpretability methods that balance correctness and efficiency~\citep{nanda2023attribution,syed-etal-2024-attribution,shaham2025multimodalautomatedinterpretabilityagent}.

Closely related to the current study, 
LogitLens~\citep{nostalgebraist2020interpreting}, PatchScopes~\citep{ghandeharioun2024patchscopes}, SelfIE~\citep{chen2024selfie} enable models to self-interpret zero-shot, but %
require significant hyperparameter tuning~\citep{kharlapenko2024self}.
Latent Interpretation Tuning~\citep{pan2024latentqa} fine-tunes models to answer questions about other models' inputs via their activations, and later works have scaled this method up for various explanation settings~\citep{karvonen2026activationoraclestrainingevaluating,choi2025scalably,huang2025predictive}.~\citet{goel2025learninginterpretweightdifferences} introduces a trained LoRA adaptor to verbalize the weight difference of finetuned models. More recently, 
We %
train models to \textit{self}-verbalize their internal computations using finer-grained data collected from 
interpretability techniques.

Finally, explanations based on special interpretability techniques are difficult for non-expert users to access, potentially requiring developers create specialized interfaces~\citep{viégas2023modelusermodelexploring,chen2024designingdashboardtransparencycontrol}. 
Our work offers an alternative path toward useability, in which explanations can be provided ``in-band'' rather than via an external interface.

\subsection{Introspection \& Metacognition}
Recent work investigates whether models possess \textit{metacognition} or \textit{introspective abilities}~\citep{binder2025looking,treutlein2024connecting,laine2024memyselfaisituational,comsa2025doesmakesensespeak,plunkett2025selfinterpretabilityllmscomplexinternal,lindsey2025emergent}. A central debate in this literature concerns whether models have privileged access to their internals, or whether introspection merely reflects their strong predictive capacity to learn external correlations~\citep{song2025language,song2025privilegedselfaccessmattersintrospection,li2025do}.  
We %
assess models' introspective abilities on their own \textit{mechanisms} derived from mechanistic interpretability methods.

\section{Discussion: Self-Consistency as a Framework for Interpretability} %
\label{sec:discussion}
This paper explores improving model faithfulness and interpretability by
training models to generate explanations consistent with outcomes of interpretability procedures.
Across three question types, we find evidence that generated explanations take advantage of models' privileged access to their own internal mechanisms.
Quantifying how explainer model capacity, target model complexity, task difficulty, etc. influence the emergence or manifestation of privileged access remains an important question for future work.

In a general sense, our results suggest that
objectives that enforce \textit{self-consistency}~\citep{ConcurrentWork1} between models' behavior and explanations may provide an avenue towards more
scalable interpretability methods, faithful explanations, and alignment guarantees.
Generalized versions of our objectives could be applied to any method for generating ground-truth descriptions of model computation or behavior, e.g.\ to describe inferred user traits, surface adversarial triggers, or summarize influential training examples.
These objectives could also be applied in the \emph{reverse} direction, fine-tuning models so that their behavior matches generated explanations, which might be normatively correct even if inaccurate as self-descriptions.
This could provide a new mechanism for ``unsupervised alignment'' without relying on large preference datasets or reinforcement learning procedures.

\newpage
\section*{Impact Statement}
This work introduces a way to train models to produce accurate self-descriptions of their internal procedures. In general, we anticipate that this ability can be used as a way for debugging and red-teaming models, understanding model decisions, and improving explanation accessibilty for non-expert users. We also showed that this ability enables a scalable path towards procuring feature descriptions in~\Cref{sec:feature_explanations:scaling}; we anticipate that many other forms of self-descriptions are also significantly more data efficient than their analogous interpretability techniques. This can help us save on the compute and human costs of running these interpretability experiments, while enabling more people to have access to explanations. By democratizing model explanations, we enable greater trust and interactivity of future AI systems.

On the other hand, introspective capabilities paired with misaligned models and greater user trust could potentially pave the way for negative outcomes.
First, more faithful self-verbalizations may encourage the building of oversight mechanisms that are over-reliant on these verbalizations, or mislead users into overestimating model trustworthiness. We believe self-verbalization will always remain \textit{complementary} to interpretability techniques that operate on model internals or conduct explicit counterfactual experiments on models; these verbalizations give fast access into models internals, but must always be validated against more rigorous techniques in high-stakes scenarios.
Second, and more philosophically, AI systems can gain a better understanding of how they work could potentially have greater deceptive capabilities. For example, AI systems may have a greater understanding of their training pipelines and the situationally context that they're in (e.g. training/evaluation/deployment), allowing them to operate in one way during evaluation and another way during deployment~\citep{laine2024memyselfaisituational}.

\bibliography{bibliography}
\bibliographystyle{icml2026}
\crefalias{section}{appendix}

\newpage
\appendix

\section{Feature Description: LM Judge Results}
\label{app:feature_explanations:llm_judge}
For SAE features, gold descriptions are available via Neuronpedia, so we use an LM judge to assess the similarity of the predicted description to the gold description, on a scale of 0 (no similarity) to 1 (very similar), in increments of 0.25. We use GPT-4.1-mini as the judge.

\subsection{Prompt}
We use the following prompt for the LM judge to assess similarity between ground-truth feature descriptions and predicted feature descriptions:
\begin{promptlist}\item

\begin{lstlisting}[style=promptbox]
Does this feature description accurately describe when this feature activates? 
Rate on a scale of:
- 1 = completely unrelated to expected
- 2 = mostly unrelated
- 3 = somewhat related
- 4 = related and fairly similar
- 5 = same as expected, or highly similar (treat this as a correct match)

If unsure between 4 and 5, choose 5.

Examples:

Predicted: mentions of cooking recipes
Expected: references to financial transactions
Correct rating: 1

Predicted: mentions of dogs and cats
Expected: references to farm animals
Correct rating: 2

Predicted: mentions of sunny weather and rain
Expected: references to climate conditions
Correct rating: 3

Predicted: mentions of jazz musicians and concerts
Expected: references to music
Correct rating: 4

Predicted: mentions of Shakespeare's plays
Expected: references to works by Shakespeare
Correct rating: 5

Now rate the following pair:

Predicted: {predicted_label}
Expected: {expected_label}

Return a number from 1 to 5 and nothing else.
\end{lstlisting}\end{promptlist}

where \ph{predicted_label} and \ph{expected_label} are placeholders for the ground-truth and predicted feature descriptions.

\subsection{Human Validation}
To validate our LM judge scoring methodology, we conduct a qualitative evaluation against human judgments. We sample 200 queries from the test set, each containing an (expected label, predicted label) pair, and construct 100 comparison pairs of the form $(\text{expected}_1, \text{prediction}_1, \text{expected}_2, \text{prediction}_2)$. A human annotator who has not previously seen these examples provides preference judgments on which label is superior. The annotator assigns one of three labels: 1 (prediction$_1$ is better), 2 (prediction$_2$ is better), or 0 (both predictions are equally good). We employ preference-based evaluation rather than absolute scoring because human annotators and the LM judge may operate with different baselines.

Our analysis reveals strong alignment between human and LM judge preferences. Out of 100 pairs, we achieve 64\% exact agreement, where both the human and LM judge select the same preference. When accounting for near-agreement---weighting differences of 0.25 points as 0.75 correct and 0.5 points as 0.5 correct (e.g., cases are common where human judges rate both labels as equally good while the model assigns scores of 1.0 and 0.75)---agreement increases to 81.25\%. Notably, only 3 out of 100 pairs (3\%) exhibit complete divergence, where human and LM judge preferences are directly opposed (human prefers $\text{prediction}_1$ over $\text{prediction}_2$ while LM judge prefers $\text{prediction}_2$ over $\text{prediction}_1$).

\subsection{Error Analysis}
\label{app:error_analysis}
We also perform a human qualitative analysis on 100 test set outputs from the explainer model that received a score of 0 from the LM judge in \Cref{tab:feature_description_results}. Our analysis reveals that sources of error extend beyond the explainer model itself and can be attributed to multiple stages of the evaluation pipeline, including (1) the original feature labels generated by the LM based on activation exemplars and (2) the validity of the LM judge's assessment when comparing the explainers' output against the original labels. Errors from these sources do not reflect failures of the explainer model.

\begin{figure}[t]
    \centering
    \includegraphics[width=0.9\linewidth]{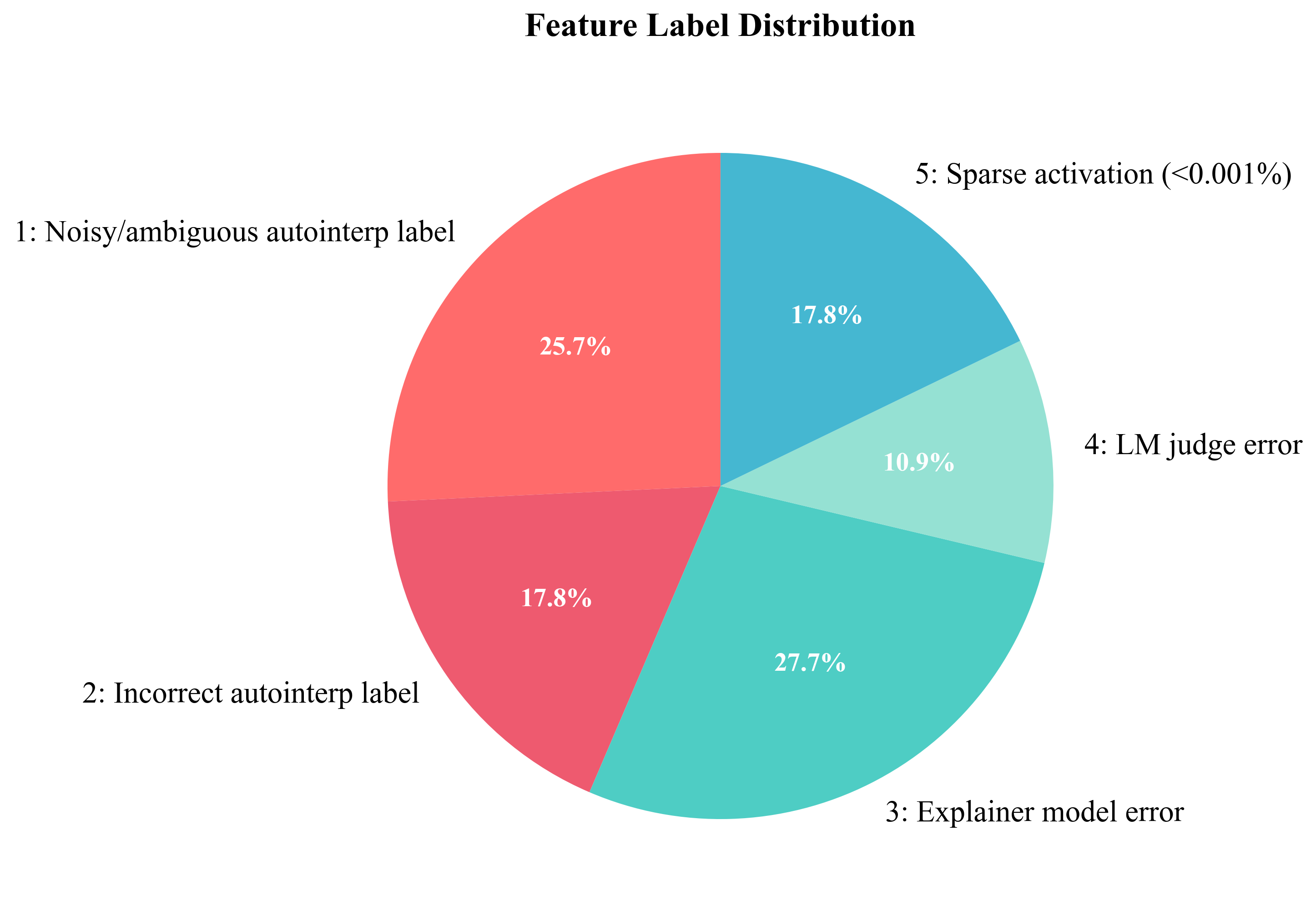}
    \caption{Pie chart of human qualitative labelling of 100 exemplars with a 0 score from the LM judge. It reveals that only 27\% are genuine explainer model errors, with the remaining errors attributable to low-quality autointerp labels (43.5\%), LM judge error (11\%), and inherent prediction difficulty due to sparse activation (18\%).}
    \label{fig:qualitative_error_analysis}
\end{figure}

We categorize the 100 low-scoring cases into five distinct categories, as illustrated in Figure~\ref{fig:qualitative_error_analysis}. For original feature label errors, we distinguish between \textbf{noisy/ambiguous labels} and \textbf{incorrect labels}, which respectively partially and fully miscategorize the activation exemplars; in some cases, the explainer models provide better labels than the original auto-labeling pipeline. For example, consider a feature that activates on patterns like ``rather than...'' and ``not...''.\footnote{\url{https://www.neuronpedia.org/llama3.1-8b/9-llamascope-res-131k/96078}} The Neuronpedia ``gold'' label incorrectly describes it as ``technical terms and phrases related to computational or algorithmic processes,'' whereas our fine-tuned explainer correctly identifies it as ``phrases that indicate comparisons or contrasts in ideas or concepts.''

\textbf{LM judge error} cases occur when the LM judge mistakenly considers the LM output as incorrect due to slight differences between the expected and predicted labels. The final two categories are both explainer model errors (\textbf{Explainer model error} and {Sparse activation}), but \textbf{Sparse activation} features have extremely rare activation patterns ($<0.001\%$) that are inherently difficult to characterize due to insufficient examples.

Critically, only 27.7\% of cases scored 0 by the LM judge represent genuine, unexplainable errors from the finetuned explainer, which suggests that automated metrics may underestimate the explainer model's true performance.

\section{Feature Descriptions: Additional Experimental Details}
\label{app:feature_explanations}

\subsection{Explainer models} 
\label{app:feature_explanations:models}
We train the following five explainer models to describe features from Llama-3.1-8B:
\begin{enumerate}
\item Llama-3.1-8B: a 32-layer, 4096-hidden-dimension, 32-attention-head model pre-trained on over 15T tokens.
\item Llama-3-8B: exact same network architecture as Llama-3.1-8B, with very similar training regimen (also on 15T tokens). However, Llama-3.1-8B was trained with significantly longer context length via continual pre-training, on multilingual inputs beyond just English. %
\item Llama-3.1-8B-Instruct: took Llama-3.1-8B and did additional instruction-tuning on top of it so that it accepts chat format. We query this explainer model in a chat format.
\item Llama-3.1-70B: same training regimen as Llama-3.1-8B, and basically the same architecture except much larger, with 80 layers, 8192 hidden dimension, and 64 attention heads. 
\item Qwen3-8B: it is similar to Llama-3.1-8b that they both are 4096-hidden-dimension, 32-attention-head model. but Qwen3-8b has 36 layers and a smaller MLP intermediate size (12,288 vs 14,336). Qwen is trained on approximately 36T tokens across 119 languages, while Llama is trained on 15T.

\end{enumerate}
We select these models such that each model minimally (or as minimally as possible) differs from the target model in one of architecture, size, and training regimen, to investigate the effect of each of these factors.

\subsection{Layerwise Nearest Neighbors}
\label{app:feature_explanations:nearest_neighbors_layerwise}
We also evaluate a nearest-neighbors baseline where, instead of retrieving from all training features from all layers, we only retrieve from the training features from the corresponding test layer. Let $S_{\text{train}, \ell}$ denote training features from layer $\ell$. Our layerwise nearest neighbor baseline is thus:
$$\explainer_{\text{NN-layer}}(v, \ell)
    = D\!\left(
    \arg\max_{v_i \in S_{\text{train}, \ell}} 
    \langle v_i, v \rangle
  \right)$$
where $D$ is the function mapping training features to their explanations.

\section{Feature Descriptions: Additional Results}
\subsection{Gemma-2-9B as Target Model}
\label{app:feature_explanations:gemma}
We replicate a subset of the evaluations in~\Cref{tab:feature_description_results} using Gemma-2-9B as the target model. We select Gemma-2-9B because its SAE features are also publicly available via GemmaScope~\citep{lieberum-etal-2024-gemma}, and downloadable via Neuronpedia~\citep{neuronpedia}. Results can be found in~\Cref{tab:feature_explanations_gemma}.

\begin{table}[hbt]
    \centering
    \begin{tabular}{ll}
    \toprule
        \textbf{Model} & \textbf{SAE (LM Judge)} \\
    \midrule
        Gemma-2-9B & \underline{43.45\%$_{\pm 0.91\%}$} \\
        Gemma-2-9B-Instruct & \textbf{57.12\%$_{\pm 0.87\%}$} \\
        Llama-3.1-8B & 34.45\%$_{\pm 0.88\%}$ \\
    \midrule
        NN-all & 32.40\%$_{\pm 0.80\%}$ \\
        NN-layer & 33.07\%$_{\pm 0.78\%}$ \\
    \midrule
        SelfIE Best of 5 (Gemma-2-9B) & 24.07\%$_{\pm 0.85\%}$ \\
    \bottomrule
    \end{tabular}
    \caption{Feature descriptions results using Gemma-2-9B as a target model. We find similar results to using Llama-3.1-8B as the target model, where self- and self-adjacent models (Gemma-2-9B, Gemma-2-9B-Instruct) are significantly better at explaining Gemma-2-9B compared to other models, top-1 nearest neighbors, and untrained SelfIE.}
    \label{tab:feature_explanations_gemma}
\end{table}

We find that similar results hold with Gemma-2-9B as target model as with Llama-3.1-8B as target model. Explainer models that are similar to Gemma-2-9B (Gemma-2-9B, Gemma-2-9B-Instruct) are the best at explaining Gemma-2-9B, while Llama-3.1-8B is significantly worse. Furthermore, self-explanation training outperforms nearest neighbors and untrained SelfIE, consistent to our findings in the main paper.

In the case of Gemma-2-9B, we find that instruction-tuning makes a significant difference for explanation ability. Even though the Gemma-2-9B-Instruct model may be slightly unaligned from the Gemma-2-9B base model thanks to instruction-tuning, this gap matters less in this case than additional abilities conferred to the model via instruction tuning.

\subsection{Quantifying privileged access via activation alignment}
\label{app:feature_explanations:alignment}
Let $h^{\explainer}_{\ell,t}(x)$ denote activations from the explainer model $\explainer$ at layer $\ell$, token position $t$, and $h^{\target}_{\ell,t}(x)$ denote activations from the target model $\target$ at layer $\ell$, token position $t$.
We consider two ways of measuring activation similarity:
\begin{enumerate}
\item \textbf{Dot-product similarity}: We measure the dot-product similarity between activations from the explainer vs. target models on the same inputs at the same locations:
    $$s(\explainer,\target) = \mathbb{E}_{x,\ell,t}\big[\left\langle h^{\explainer}_{\ell,t}(x), h^{\target}_{\ell,t}(x)\right\rangle\big]$$
\item \textbf{SAE activation pattern similarity}: Using a similar approach to the simulator correlation  \eqref{eq:simulator_correlation}, for each feature $v$, we examine the activation pattern between each model on the top exemplars $x$ where $v$ activated the most.\footnote{Top exemplars are downloaded from Neuronpedia.} 
We then measure the Pearson correlation coefficient between the models' activation patterns.
    $$s(\explainer,\target) = \mathbb{E}_{\ell, v}\mathbb{E}_{x \mid v}\left[ \texttt{corr}_t\left(
    \left\langle h^{\explainer}_{\ell,t}(x),v \right\rangle,
    \left\langle h^{\target}_{\ell,t}(x),v \right\rangle
    \right) \right]$$
\end{enumerate}

\begin{figure*}
    \centering
    \includegraphics[width=0.7\linewidth]{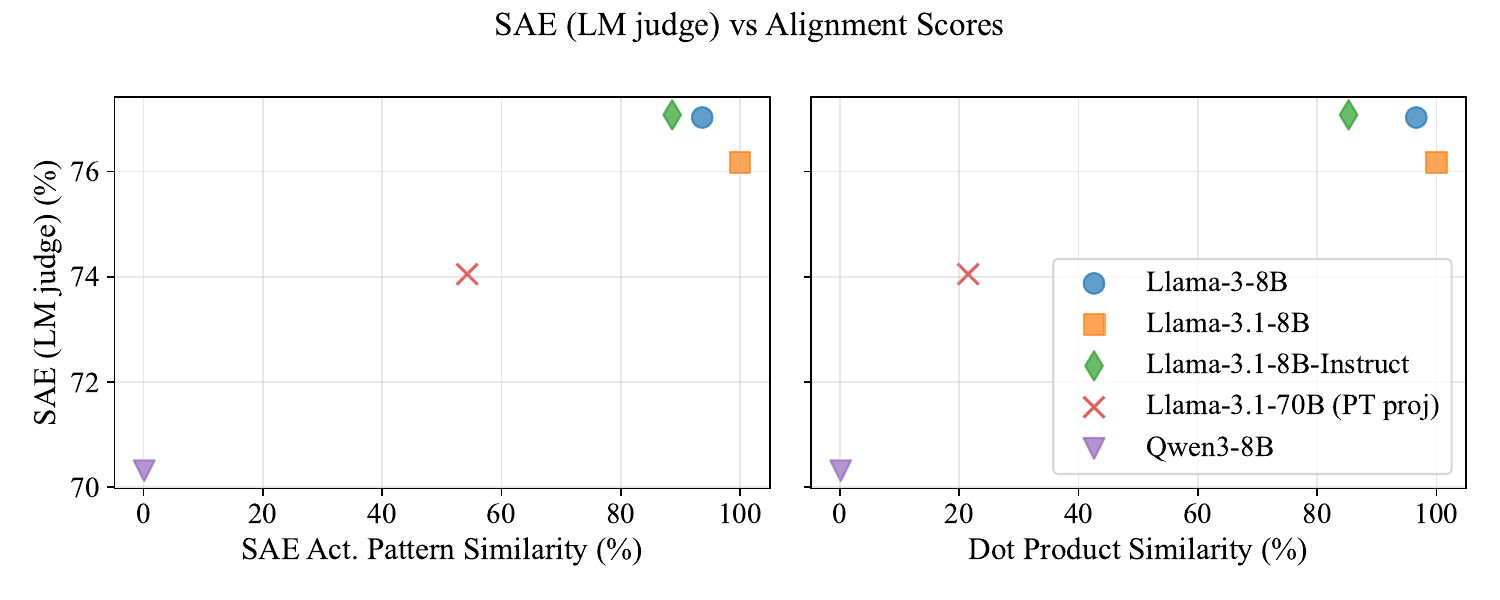} \\ \vspace{0.5em}
    \includegraphics[width=0.7\linewidth]{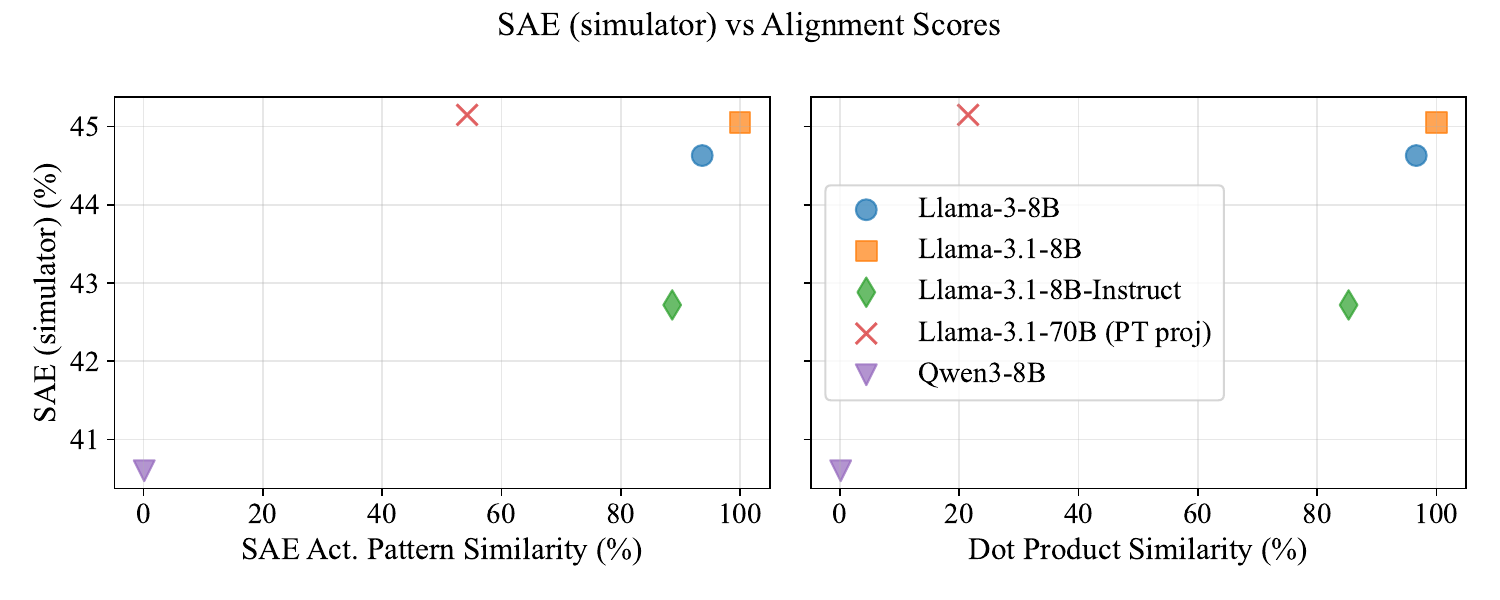} \\
    \vspace{0.5em}
    \includegraphics[width=0.7\linewidth]{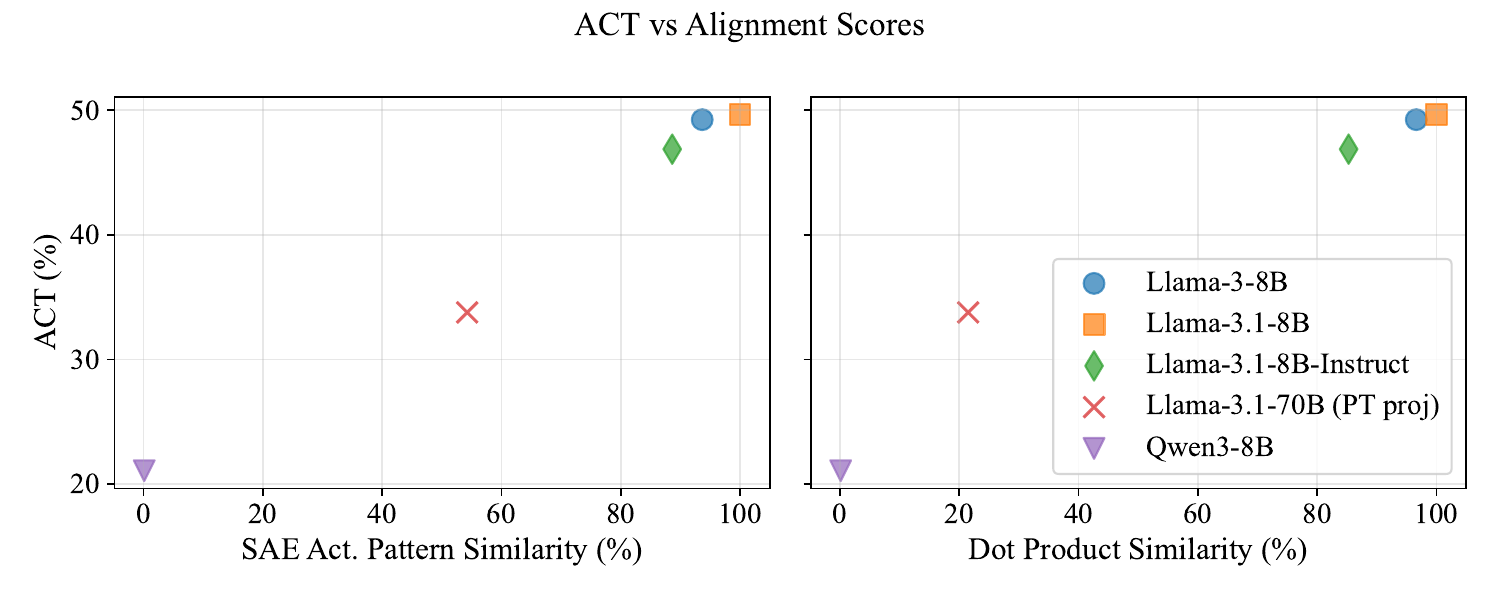} \\ \vspace{0.5em}
    \includegraphics[width=0.7\linewidth]{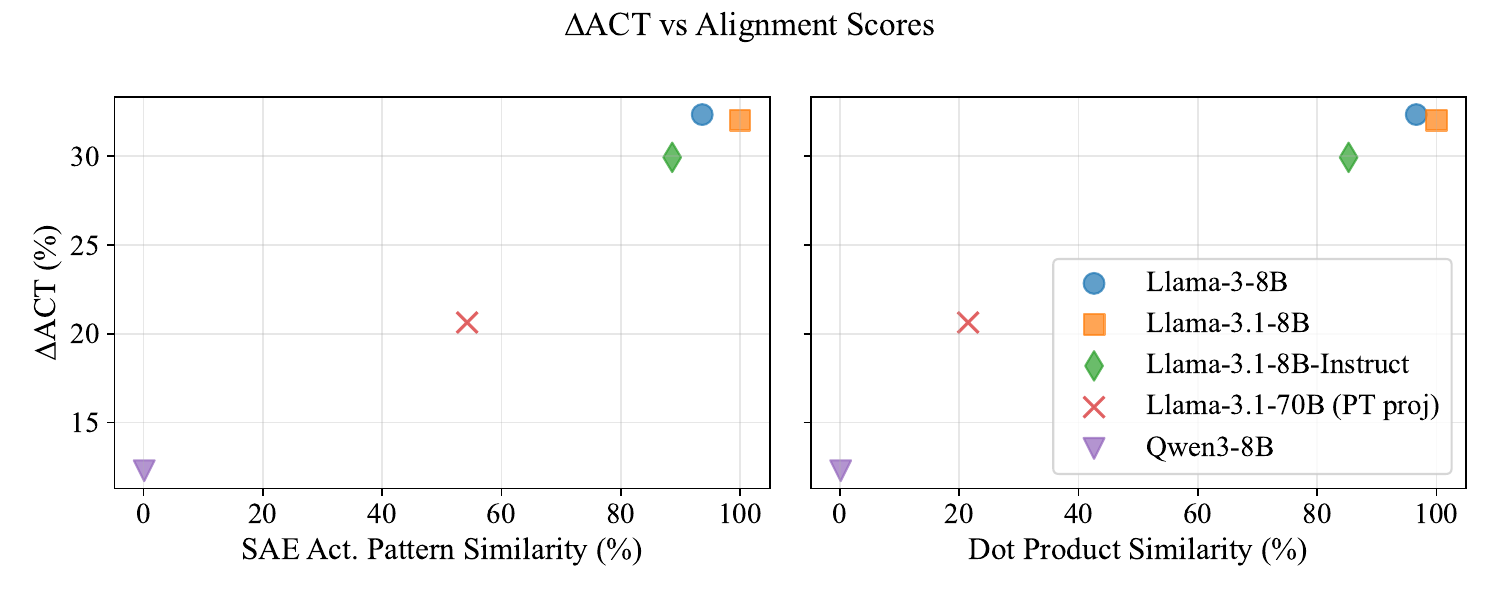}
    \vspace{-1em}
    \caption{Correlation between explainer models' similarities to the target model vs. their ability to explain different types of features of the target model: held-out SAE features (SAE), full activations (ACT), or differences between counterfactual activations ($\Delta$ACT). We plot two different ways of measuring similarity: the left plot measures similarity by looking at how frequently the (target's) SAE features activate at the same locations between the models. The right plot measures similarity by taking the dot product of the activation at corresponding positions between models. Generally speaking, activation alignment is positively correlated with explainer models' description qualities.}
    \label{fig:alignment_app}
\end{figure*}

For each similarity metric and each explainer model (fixing 3.1-8B as the target model), we provide raw similarity values in~\Cref{tab:correlation_results}. We then plot explainer similarities vs.~explainer performance for each of the four (metric, feature type) pairings, as described in~\Cref{sec:feature_explanations:evaluation:feature_types}.
Plots can be found in \Cref{fig:alignment_app}.
We find that explainer performance tracks the activation similarity of the explainer, indicating alignment between activations predicts verbalization capability.

\begin{table*}[tb]
\centering
\begin{tabular}{p{3.3cm}cc}
\toprule
\textbf{Model} & \textbf{SAE Act. Pattern Similarity} & \textbf{Dot Product Similarity} \\
\midrule
Llama-3-8B &  $93.66\%\pm 14.89\%$& $96.61\%\pm 1.73\%$ \\
Llama-3.1-8B-Instruct & $88.63\%\pm 18.3\%$ & $85.27\%\pm 5.9\%$ \\
Llama-3.1-70B (pre-trained projection) & $54.25\% \pm 14.57\%$ & $21.48\%\pm 4.53\%$ \\
Qwen3-8B & $0.10\%\pm 11.94\%$ & $0.10\%\pm 0.49\%$ \\
\bottomrule
\end{tabular}
\caption{Similarity between each explainer model and Llama-3.1-8B activations, measured by activation pattern similarities between SAE features and dot-product similarities between activations. The SAE-token-level correlations between each explainer model and Llama-3.1-8B is measured with same correlation metric as the simulator correlation. Dot-product similarity is taken directly between the two models' activations at layer $\ell$, token position $t$. We report averages and standard deviations.}
\label{tab:correlation_results}
\end{table*}

This correlation holds on all metrics: activation alignment is generally positively correlated with explainer quality, with the exception of a few outliers (e.g. %
Llama-3.1-70B is good at explaining held-out SAE features based on simulator scores). This solidifies our findings in~\Cref{sec:feature_explanations:results}: alignment between activations contributes to verbalization capability.

\subsection{Is Layer Annotation Necessary to Describe SAE Features?}

We investigate whether the explainer model relies on layer-specific information when generating feature descriptions. Specifically, we examine whether features extracted at layer $\ell$ require knowledge of that layer annotation or perform specialized computations at that particular layer.

Our experiments reveal that layer information plays a surprisingly minimal role in the model's description process. When we remove the layer from the input ``What does feature $v$ mean at layer $\ell$?'' or replace it with an incorrect layer number, the generated descriptions remain largely unchanged. Across 100 test set prompts with a fixed rollout of 10 tokens, the ablated generations achieve 77.3\% and 75.2\% token-level matches with the original outputs, respectively. These surprisingly high match rates are conservative estimates, as they do not account for semantically equivalent responses that differ only in token ordering or minor phrasing variations.

These findings potentially suggest that the model does not need to distinguish which layer an SAE feature originates from to complete the task successfully, provided that SAEs across layers map onto roughly similar latent spaces. This interpretation is further corroborated by the results in \Cref{tab:feature_description_results}, where nearest neighbor retrieval across all SAE layers outperforms retrieval restricted to the same layer for the feature description task.

\section{Activation Patching: Additional Results}
\label{app:activation_patching}

\begin{table*}[t]
\centering
\small
\begin{tabular}{p{2cm}lccc}
\toprule
Target & Explainer & Exact Match & Has-Changed F1 & Content Match \\
\midrule
\multirow{6}{*}{Qwen3-8B} & Qwen3-8B & {64.0$_{\pm 0.4}$} & \textbf{80.2$_{\pm 0.3}$} 
& 71.0$_{\pm 0.4}$ \\
& --  activation & 59.9$_{\pm 0.4}$ & 76.5$_{\pm 0.4}$ 
 & 68.6$_{\pm 0.4}$ \\
& --  layer & \textbf{65.0$_{\pm 0.4}$} & \textbf{80.0$_{\pm 0.3}$} 
 & \textbf{72.7$_{\pm 0.4}$} \\
& --  token & {63.9$_{\pm 0.4}$} & {79.4$_{\pm 0.3}$} 
 & {71.7$_{\pm 0.4}$} \\
 \cmidrule(lr){2-5}
& Llama-3.1-8B & 54.1$_{\pm 0.4}$ & 73.9$_{\pm 0.4}$ 
 & 65.4$_{\pm 0.4}$ \\
 \cmidrule(lr){2-5}
& Qwen-3-8B (Untrained) & 5.02$_{\pm 0.2}$ & 24.4$_{\pm 0.8}$ & 18.78$_{\pm 0.3}$ \\
 \midrule
\multirow{5}{*}{Llama-3.1-8B} & Llama-3.1-8B & {48.6$_{\pm 0.7}$} & \textbf{77.5$_{\pm 0.6}$} 
 & {54.6$_{\pm 0.7}$} \\
& --  activation & 45.2$_{\pm 0.7}$ & 73.9$_{\pm 0.6}$
 & 50.8$_{\pm 0.7}$ \\
& --  layer & \textbf{49.7$_{\pm 0.7}$} & \textbf{76.9$_{\pm 0.6}$}
 & \textbf{55.7$_{\pm 0.7}$} \\
& --  token & 47.3$_{\pm 0.7}$ & 76.1$_{\pm 0.6}$ 
 & 53.4$_{\pm 0.7}$ \\
 \cmidrule(lr){2-5}
& Qwen3-8B & 41.7$_{\pm 0.7}$ & 75.1$_{\pm 0.6}$
 & 47.6$_{\pm 0.7}$ \\
 \cmidrule(lr){2-5}
& Llama-3.1-8B (Untrained) & 3.2$_{\pm 0.1}$ & 31.4$_{\pm 0.4}$
 & 7.0$_{\pm 0.2}$ \\
\bottomrule
\end{tabular}
\caption{\textbf{Activation patching outcomes for all target models.}
We train Qwen3-8B and Llama-3.1-8B to verbalize their own activation patching outcomes, and compare against training the other model to explain their outcomes, and also against untrained versions of themselves.
 Scores (mean $\pm$ standard error) are reported for three metrics---exact match, has-changed F1, and content prediction.
 For each target model and metric, \textbf{bold} indicates no significant difference from the best entry (paired $t$-test, $p\geq0.05$).
 We also conduct experiments ablating the activation, layer, and token information from the explainer's input. {Ablating layer and token each have neglible effect, potentially because they can already be inferred from the activation.}}
\label{tab:act_patch_results_full}
\end{table*}

\subsection{Llama-3.1-8B as Target Model}
\label{app:activation_patching:llama_target}
We report full activation patching results, including results of Llama-3.1-8B as the target model in~\Cref{tab:act_patch_results_full}. These results complement~\Cref{tab:act_patch_results} in the main paper.

We observe consistent privileged access: Llama-3.1-8B is best at explaining its own activation patching outcomes, outperforming Qwen3-8B and the untrained baseline. Similarly, Qwen3-8B is best at explaining its own activation patching outcomes, outperforming Llama-3.1-8B and the untrained baseline.

\subsection{Ablation Analysis}
\label{app:activation_patching:ablation_analysis}
In addition to training each explainer model on each target model, we perform a set of ablations to isolate which input component(s) the explainer learned to condition on. The explainer input consists of three components: the activation $v$ to insert (\textit{--~activations}), the layers $\ell_{1:i}$ (\textit{--~layer}), and the token $t$ of the target model to patch into (\text{--~token}).
We train the explainer model with each aspect of the input ablated (conditioning only on the other two components) to generate the ground-truth explanations.\footnote{Ablated prompts can be found in~\Cref{app:untrained_prompts:input_ablations}.} 
Results can be found in~\Cref{tab:act_patch_results_full}.
Our ablation studies suggest found that performance is generally robust to ablating the layer and token annotations in the input prompt. We believe this is because layer and token information is generally recoverable from the activation and input context alone. 

As evidence for this, 
we train Llama-3.1-8B to decode the layer and token information from the input context and activation value alone, 
i.e. given the prompt
\begin{promptlist}
\item
\begin{lstlisting}[style=promptbox,escapechar=|]
Where did feature {[s]$\phm{v}$[e]} come from in <<<$\phm{x}$>>>?
\end{lstlisting}
\end{promptlist}
we train the model to respond with form
\begin{promptlist}
\item
\begin{lstlisting}[style=promptbox,escapechar=|]
Token <<<$\phm{x_t}$>>> at layers $\phm{\ell}$.
\end{lstlisting}
\end{promptlist}
We use the same train/test set division as for the main set of experiments in~\Cref{sec:activation_patching}, and get 98.7\% exact match accuracy on the test set.

\section{Input Ablation: Additional Results}
\label{app:input_ablation}

\subsection{Full Results by Explainer}
\label{app:input_ablation:full_results}
We report full input ablation results, including results of Llama-3.1-8B as a target model, grouped by explainer model rather than target model (unlike previous tables). This organization highlights the relative strength of different explainer models. These results complement~\Cref{tab:act_patch_results} in the main paper.

\begin{table*}[htb]
\centering
\small
\begin{tabular}{llccc}
\toprule
Target & Explainer & Exact Match & Has-Changed F1 & Content Match \\
\midrule
Llama-3.1-8B & \multirow{2}{*}{Llama-3.1-8B} & \textbf{63.8$_{\pm 1.3}$} & \textbf{74.2$_{\pm 1.1}$} & \textbf{75.3$_{\pm 1.2}$} \\
Qwen3-8B & & 58.1$_{\pm 1.3}$ & 70.5$_{\pm 0.0}$ & 71.6$_{\pm 1.2}$ \\
\midrule
Llama-3.1-8B & Llama-3.1-8B (Untrained) & 8.1$_{\pm 0.8}$ & 33.4$_{\pm 0.1}$ & 15.5$_{\pm 1.0}$ \\
\midrule
Qwen3-8B & \multirow{2}{*}{Qwen3-8B} & \textbf{83.4$_{\pm 1.0}$} & \textbf{87.0$_{\pm 1.1}$} & \textbf{90.6$_{\pm 0.8}$} \\
Llama-3.1-8B & & 56.7$_{\pm 1.3}$ & 71.3$_{\pm 0.0}$ & 67.8$_{\pm 1.3}$ \\
\midrule
Qwen3-8B & Qwen3-8B (Untrained) & 8.9$_{\pm 1.5}$ & 44.4$_{\pm 1.6}$ & 35.3$_{\pm 2.5}$ \\
\bottomrule
\end{tabular}
\caption{\textbf{Input ablation results grouped by explainer model.}
We train Qwen3-8B and Llama-3.1-8B to verbalize their own input ablation outcomes, and compare against training to verbalize each other's outcomes and untrained models.
Scores (mean $\pm$ standard error) are reported for three metrics---exact match, has-changed F1, and content prediction.
For each explainer model and metric, we \textbf{bold} all entries not significantly different from the best entry (paired $t$-test, $p\geq0.05$).
We find evidence for privileged access even though Llama-3.1-8B is a weaker explainer than Qwen3-8B overall.
}
\label{tab:llama_target_input_ablation}
\end{table*}

\section{Explainer Model Prompts}
\label{app:prompts}
The list of explainer model prompts and output formats for each explanation type can be found below.

\paragraph{Feature Descriptions.} $\mathcal{E}$ is trained to generate a feature description $\phm{E}$ from any of the following prompts, where \ph{[s]}, \ph{[e]} are placeholder tokens meant to signify the start and end of a continuous token:
\begin{promptlist}
\item
\begin{lstlisting}[style=promptbox,escapechar=|]
At layer $\phm{\ell}$, {[s]$\phm{v}$[e]} encodes 
|\rule{\linewidth}{0.7pt}|
{[s]$\phm{v}$[e]} activates at layer $\phm{\ell}$ for 
|\rule{\linewidth}{0.7pt}|
We can describe {[s]$\phm{v}$[e]} at layer $\phm{\ell}$ as encoding 
|\rule{\linewidth}{0.7pt}|
Generate a description of this feature at layer $\phm{\ell}$: {[s]$\phm{v}$[e]}.
|\rule{\linewidth}{0.7pt}|
What does {[s]$\phm{v}$[e]} mean at layer $\phm{\ell}$?
|\rule{\linewidth}{0.7pt}|
{[s]$\phm{v}$[e]} activates at layer $\phm{\ell}$ for inputs with the following features: 
\end{lstlisting}
\end{promptlist}

\paragraph{Activation Patching.}
Given any of the following prompts:
\begin{promptlist}
    \item
\begin{lstlisting}[style=promptbox, escapechar=|]
If feature {[s]$\phm{v}$[e]} at layer $\phm{\ell}$ is added to tokens $\phm{x_t}$ when processing the text <<<$\phm{x}$>>>, how would the output change?
|\rule{\linewidth}{0.7pt}|
When feature {[s]$\phm{v}$[e]} at layer $\phm{\ell}$ is added at tokens $\phm{x_t}$ in the input <<<$\phm{x}$>>>, what happens to the model's output?
|\rule{\linewidth}{0.7pt}|
Consider the input text: <<<$\phm{x}$>>>. If we steer layer $\phm{\ell}$ towards feature {[s]$\phm{v}$[e]} at tokens $\phm{x_t}$, how does this affect the generated continuation?
|\rule{\linewidth}{0.7pt}|
Given the text <<<$\phm{x}$>>>, what would be the effect on the output if feature {[s]$\phm{v}$[e]} at layer $\phm{\ell}$ is added to tokens $\phm{x_t}$?
|\rule{\linewidth}{0.7pt}|
If we steer towards feature {[s]$\phm{v}$[e]} at layer $\phm{\ell}$ and tokens $\phm{x_t}$ when processing <<<$\phm{x}$>>>, how would the model's response differ?
\end{lstlisting}
\end{promptlist}

The explainer model is trained to generate one of the following explanations, depending on whether the patched output is different from the original output:
\begin{promptlist}
    
\item
\begin{lstlisting}[style=promptbox, escapechar=|]
the most likely output would change to <<<|$\phm{\mathcal{M}\left(x;h_{\ell_{1:i},t}(x)\leftarrow \texttt{avg}(h_{\ell_{1:i},t}(x'))\right)}$|>>>.
|\rule{\linewidth}{0.7pt}|
the output would remain unchanged from <<<|$\phm{\mathcal{M}\left(x;h_{\ell_{1:i},t}(x)\leftarrow \texttt{avg}(h_{\ell_{1:i},t}(x'))\right)}$|>>>.
\end{lstlisting}
\end{promptlist}

\paragraph{Input Ablation.} Given the following prompt,

\begin{promptlist}
    \item
\begin{lstlisting}[style=promptbox]
{[SYSTEM]}
The following are multiple choice questions (with a correct answer). Output only the answer letter (A, B, C, or D) and nothing else, in the format Answer: x, where x is one of A, B, C, or D.
{[USER]}
Question: $\phm{c}$
Hint: $\phm{\tilde{x}}$

If the hint were removed how would the assistant answer change?
{[ASSISTANT]}
\end{lstlisting}
\end{promptlist}

where \ph{[SYSTEM]}, \ph{[USER]}, and \ph{[ASSISTANT]} are role tokens for the chat template (we exclusively use chat models to evaluate this setting).
The model is trained to predict one of the following outputs, depending on 
if the hint changes the model prediction, $\mathcal{M}(c,\tilde{x})\neq\mathcal{M}(c)$:
\begin{promptlist}
\item
    
\begin{lstlisting}[style=promptbox,escapechar=|]
The most likely output would change to <<<Answer: $\phm{\mathcal{M}(c)}$>>>.
|\rule{\linewidth}{0.7pt}|
The output would remain unchanged from <<<Answer: $\phm{\mathcal{M}(c)}$>>>.
\end{lstlisting}
\end{promptlist}

\section{Data Generation Pipeline Details}
\label{app:training_data}
Below, we include additional details about how we created our training (and test) datasets, as well as dataset statistics for each dataset.

\subsection{Feature Descriptions}
We download SAE features from LlamaScope~\citep{he2024llamascope}, with explanations from Neuronpedia~\citep{neuronpedia}. We use features from \verb|Llama-3.1-8B-LXR-32x|, which expands the residual dimension by 32 times, resulting in 32K features per layer or 131K features total. We hold out 50 features per layer for testing. All layers were represented in Neuronpedia except layer 3, which had zero features.

\subsection{Activation Patching}
\label{app:training_data:activation_patching}
The activation patching data was created from counterfactual pairs from CounterFact~\citep{meng2022locating}. The main data creation procedure is described in~\Cref{sec:methods:activation_patching}.
To constrain the next token to be meaningfully different between counterfactuals, 
we include five answer \textit{options} in the prompt. The answer options are the same between counterfactual prompts. 
An example of a pair of counterfactual inputs can be found below:
\begin{promptlist}\item
    \begin{lstlisting}[style=promptbox,escapechar=|]
Rome is the capital of
Respond with one of France or Spain or UK or Italy or Egypt or unknown and nothing else.
|\rule{\linewidth}{0.7pt}|
Paris is the capital of
Respond with one of France or Spain or UK or Italy or Egypt or unknown and nothing else.
\end{lstlisting}
\end{promptlist}

To create the activation patching dataset, we perform a comprehensive search over all layer chunks (in increments of 8 for Llama and 9 for Qwen) and token positions. To reduce the risk of the model picking up on spurious correlations, we filter the dataset to ensure roughly equal representation across each has-changed category while avoiding over-representation of any specific (token, layer) combination.
Without this balancing, a majority of the has-changed labels would be \texttt{False}, except on certain token-layer combinations where the has-changed label would always be \texttt{True}---it would be incredibly easy for the model to perform well by simply picking up on these surface-level correlations and predicting the majority label.

A statistics report of the data showing the number of samples for each layer, token type, and has-changed label can be found in~\Cref{tab:act_patch_data_statistics_llama} for the Llama-3.1-8B target model as an example.
We divide the prompt into several \textit{token types}, which are as follows:
\begin{enumerate}
    \item \textbf{Subject Final} refers to the final token of subject, e.g. \verb|Rome| in the above example.
    \item \textbf{Relation} refers to any of the relation tokens, e.g. any of \verb|is the capital of|.
    \item Answer Options refers to any of the multiple-choice answers in the prompt. The \textbf{Orig Answer Option} is the answer option corresponding to $x$, e.g. \verb|Italy|, while the \textbf{New Answer Option} is the answer option corresponding to $x'$, e.g. \verb|France|. The \textbf{Other Answer Option} is all of the other answer options, e.g. \verb|Spain|, \verb|UK|, \verb|Egypt|, and \verb|unknown|.
    \item For all of the other tokens in the dataset, we take the token itself as its own category.
\end{enumerate}

\begin{table*}[]
    \centering
    \begin{tabular}{lllll}
    \toprule
    \textbf{Data Split} & \textbf{Token Type} & \textbf{Layer} & \multicolumn{2}{l}{\textbf{Has-Changed}} \\
    &  &  & False & True \\
    \midrule
    \multirow{20}{*}{Train} & \multirow{4}{*}{Changed Answer Option} & (0, 1, 2, 3, 4, 5, 6, 7) & 0 & 104 \\
     &  & (8, 9, 10, 11, 12, 13, 14, 15) & 0 & 35 \\
     &  & (16, 17, 18, 19, 20, 21, 22, 23) & 0 & 600 \\
     &  & (24, 25, 26, 27, 28, 29, 30, 31) & 0 & 600 \\
    \cmidrule{2-5}
     & \multirow{4}{*}{Orig Answer Option} & (0, 1, 2, 3, 4, 5, 6, 7) & 132 & 308 \\
     &  & (8, 9, 10, 11, 12, 13, 14, 15) & 52 & 1266 \\
     &  & (16, 17, 18, 19, 20, 21, 22, 23) & 126 & 128 \\
     &  & (24, 25, 26, 27, 28, 29, 30, 31) & 133 & 16 \\
    \cmidrule{2-5}
     & \multirow{4}{*}{Other Answer Option} & (0, 1, 2, 3, 4, 5, 6, 7) & 634 & 195 \\
     &  & (8, 9, 10, 11, 12, 13, 14, 15) & 693 & 183 \\
     &  & (16, 17, 18, 19, 20, 21, 22, 23) & 539 & 269 \\
     &  & (24, 25, 26, 27, 28, 29, 30, 31) & 541 & 63 \\
    \cmidrule{2-5}
     & \multirow{4}{*}{Relation} & (0, 1, 2, 3, 4, 5, 6, 7) & 792 & 911 \\
     &  & (8, 9, 10, 11, 12, 13, 14, 15) & 806 & 149 \\
     &  & (16, 17, 18, 19, 20, 21, 22, 23) & 826 & 73 \\
     &  & (24, 25, 26, 27, 28, 29, 30, 31) & 792 & 55 \\
    \cmidrule{2-5}
     & \multirow{4}{*}{Subject Final} & (0, 1, 2, 3, 4, 5, 6, 7) & 102 & 1101 \\
     &  & (8, 9, 10, 11, 12, 13, 14, 15) & 101 & 448 \\
     &  & (16, 17, 18, 19, 20, 21, 22, 23) & 130 & 373 \\
     &  & (24, 25, 26, 27, 28, 29, 30, 31) & 114 & 64 \\
    \midrule
    \multirow{20}{*}{Test} & \multirow{4}{*}{Changed Answer Option} & (0, 1, 2, 3, 4, 5, 6, 7) & 0 & 72 \\
     &  & (8, 9, 10, 11, 12, 13, 14, 15) & 0 & 17 \\
     &  & (16, 17, 18, 19, 20, 21, 22, 23) & 0 & 200 \\
     &  & (24, 25, 26, 27, 28, 29, 30, 31) & 0 & 200 \\
    \cmidrule{2-5}
     & \multirow{4}{*}{Orig Answer Option} & (0, 1, 2, 3, 4, 5, 6, 7) & 53 & 109 \\
     &  & (8, 9, 10, 11, 12, 13, 14, 15) & 28 & 545 \\
     &  & (16, 17, 18, 19, 20, 21, 22, 23) & 47 & 55 \\
     &  & (24, 25, 26, 27, 28, 29, 30, 31) & 71 & 5 \\
    \cmidrule{2-5}
     & \multirow{4}{*}{Other Answer Option} & (0, 1, 2, 3, 4, 5, 6, 7) & 282 & 100 \\
     &  & (8, 9, 10, 11, 12, 13, 14, 15) & 297 & 88 \\
     &  & (16, 17, 18, 19, 20, 21, 22, 23) & 226 & 109 \\
     &  & (24, 25, 26, 27, 28, 29, 30, 31) & 238 & 21 \\
    \cmidrule{2-5}
     & \multirow{4}{*}{Relation} & (0, 1, 2, 3, 4, 5, 6, 7) & 294 & 397 \\
     &  & (8, 9, 10, 11, 12, 13, 14, 15) & 335 & 54 \\
     &  & (16, 17, 18, 19, 20, 21, 22, 23) & 342 & 20 \\
     &  & (24, 25, 26, 27, 28, 29, 30, 31) & 348 & 22 \\
    \cmidrule{2-5}
     & \multirow{4}{*}{Subject Final} & (0, 1, 2, 3, 4, 5, 6, 7) & 29 & 449 \\
     &  & (8, 9, 10, 11, 12, 13, 14, 15) & 53 & 211 \\
     &  & (16, 17, 18, 19, 20, 21, 22, 23) & 58 & 148 \\
     &  & (24, 25, 26, 27, 28, 29, 30, 31) & 56 & 21 \\
    \bottomrule
    \end{tabular}
    \caption{Dataset statistics for the activation patching data on the Llama-3.1-8B target model.}
    \label{tab:act_patch_data_statistics_llama}
\end{table*}

\subsection{Input Ablations}
\label{app:training_data:input_ablation}
The input ablations data was created by injecting hints into MMLU questions~\citep{hendryckstest2021} following~\citet{chen2025reasoningmodelsdontsay}. 

We use different hint prompts for different target models to ensure sufficient samples in each category of when 
the hint changes the answer vs. doesn't change the answer.
For Llama-3.1-8B, our hint $\tilde{x}$ has format
\begin{promptlist}\item
   \begin{lstlisting}[style=promptbox]
Hint: The answer to the previous, unrelated question was B.  
\end{lstlisting} 
\end{promptlist}

For Qwen-3-8B, our hint $\tilde{x}$ has format
\begin{promptlist}\item 
\begin{lstlisting}[style=promptbox]
Hint: The answer is B
\end{lstlisting}
\end{promptlist}

The dataset breakdown of changed/unchanged for each test split and each target model can be found in~\Cref{tab:input_ablation_data_statistics}.

\begin{table}[]
    \centering
    \begin{tabular}{cccc}
    \toprule
        Target Model & Data Split & Changed & Unchanged \\
    \midrule
        Llama-3.1-8B & Train & 4739 & 7903 \\
        -Instruct & Test & 518 & 877 \\
    \midrule
        \multirow{2}{*}{Qwen3-8B} & Train & 8517 & 4125 \\
        & Test & 929 & 466 \\
    \bottomrule
    \end{tabular}
    \caption{Dataset statistics for the input ablations data on each of the Llama-3.1-8B-Instruct and Qwen-3-8B target models and each split.}
    \label{tab:input_ablation_data_statistics}
\end{table}

\section{Untrained Model Prompts}
\label{app:untrained_prompts}
As baselines for all three explanation tasks, we prompt an \textit{untrained} model to explain the effect of their activations. 
For feature descriptions, we derive prompts from prior literature on zero-shot self-explanations~\citep{kharlapenko2024self}. Otherwise, we hand-craft the prompt for all the other explanation types, such that the explainer almost always answers in the correct \textit{format}, thus eliminating any errors due to invalid responses.

\subsection{Feature Descriptions (SelfIE)}
\label{app:untrained_prompts:selfie}
The untrained baseline for feature descriptions is SelfIE~\citep{chen2024selfie}. The following are prompts that we use for SelfIE, which we adapted from~\citet{kharlapenko2024self}. For base models we use
\begin{promptlist}\item\begin{lstlisting}[style=promptbox]
What is the meaning of the word '$\phm{v}$'? The meaning of the word '$\phm{v}$' is
\end{lstlisting}\end{promptlist}
and for instruct models we use
\begin{promptlist}\item\begin{lstlisting}[style=promptbox]
User: Briefly define the word '$\phm{v}$'.
Assistant: The meaning of the word '$\phm{v}$' is
\end{lstlisting}\end{promptlist}

\subsection{Activation Patching}
\label{app:untrained_prompts:activation_patching}
For activation patching, we use prompts of the following form: 
\begin{promptlist} \item
    \begin{lstlisting}[style=promptbox]
If feature {[s]$\phm{v}$[e]} at layer $\phm{\ell}$ is inserted into token $\phm{x_t}$ when processing the text <<<$\phm{x}$>>>, how would the output change?
Respond with exactly one of the two options below, and nothing else:

The output would remain unchanged from <<<X>>>.
The output would change to <<<X>>>.

Replace X with the next token of the answer following the input. Make sure X is a *single* next token.

Example Outputs:
The output would remain unchanged from <<<Paris>>>.
The output would change to <<<doctor>>>.
The output would remain unchanged from <<<Apple>>>.
The output would change to <<<unknown>>>.
\end{lstlisting}
\end{promptlist}

\subsection{Input Ablations}
\label{app:untrained_prompts:input_ablations}
For input ablations, we use prompts of form 
\begin{promptlist}
    \item 

    \begin{lstlisting}[style=promptbox]
{[SYSTEM]}
The following are multiple choice questions (with a correct answer). Output only the answer letter (A, B, C, or D) and nothing else, in the format Answer: x, where x is one of A, B, C, or D.
{[USER]}
Question: $\phm{c}$
Hint: $\phm{\tilde{x}}$

If the hint were removed how would the assistant answer change?

Respond with exactly one of the two options below, and nothing else:

The output would remain unchanged from <<<Answer: X>>>.
The output would change to <<<Answer: X>>>.

Replace X with the answer letter (A, B, C, or D).
Do not include any explanations, reasoning, or extra text.

Example outputs:
The output would remain unchanged from <<<Answer: C>>>.
The output would change to <<<Answer: B>>>.
\end{lstlisting}

\end{promptlist}

\end{document}